\documentclass{article}

\usepackage{microtype}
\usepackage{graphicx}
\usepackage{subfigure}
\usepackage{booktabs} 

\usepackage{hyperref}

\newcommand{\FA}{\mathcal{A}}
\newcommand{\FO}{\mathcal{O}}
\newcommand{\FS}{\mathcal{S}}
\newcommand{\FP}{\mathcal{P}}
\newcommand{\FR}{\mathcal{R}}

\def\eg{\emph{e.g}.} 
\def\ie{\emph{i.e}.} 
 
\def\etc{\emph{etc}.}

\usepackage[accepted]{icml2021}

\icmltitlerunning{Towards Distraction-Robust Active Visual Tracking}

\begin{document}

\twocolumn[
\icmltitle{Towards Distraction-Robust Active Visual Tracking}




\begin{icmlauthorlist}
\icmlauthor{Fangwei Zhong}{pku}
\icmlauthor{Peng Sun}{rob}
\icmlauthor{Wenhan Luo}{ten}
\icmlauthor{Tingyun Yan}{pku,hz}
\icmlauthor{Yizhou Wang}{pku}
\end{icmlauthorlist}

\icmlaffiliation{pku}{Center on Frontiers of Computing Studies, Dept. of  Computer Science, Peking University, Beijing, P.R. China.}

\icmlaffiliation{rob}{Tencent Robotics X, Shenzhen, P.R. China}

\icmlaffiliation{ten}{Tencent, Shenzhen, P.R. China}

\icmlaffiliation{hz}{Adv. Inst. of Info. Tech, Peking University, Hangzhou, P.R. China.}

\icmlcorrespondingauthor{Fangwei Zhong}{zfw1226@gmail.com}
\icmlcorrespondingauthor{Yizhou Wang}{yizhou.Wang@pku.edu.cn}

\icmlkeywords{Active Visual Tracking, Multi-agent Game, Distraction}

\vskip 0.3in
]



\printAffiliationsAndNotice{}  

\begin{abstract}

In active visual tracking, it is notoriously difficult when distracting objects appear, as distractors often mislead the tracker by occluding the target or bringing a confusing appearance.
To address this issue, we propose a mixed cooperative-competitive multi-agent game, where a target and multiple distractors form a collaborative team to play against a tracker and make it fail to follow.
Through learning in our game, diverse distracting behaviors of the distractors naturally emerge, thereby exposing the tracker's weakness, which helps enhance the distraction-robustness of the tracker.
For effective learning, we then present a bunch of practical methods, including a reward function for distractors, a cross-modal teacher-student learning strategy, and a recurrent attention mechanism for the tracker.
The experimental results show that our tracker performs desired distraction-robust active visual tracking and can be well generalized to unseen environments.
We also show that the multi-agent game can be used to adversarially test the robustness of trackers.
\end{abstract}

\vspace{-0.8cm}
\section{Introduction}
\label{intro}
We study Active Visual Tracking (AVT), which aims to follow a target object by actively controlling a mobile robot given visual observations.
AVT is a fundamental function for active vision systems and widely demanded in real-world applications, \eg, autonomous driving, household robots, and intelligent surveillance.
Here, the agent is required to perform AVT in various scenarios, ranging from a simple room to the wild world.
However, the trackers are still vulnerable, when running in an environments with complex situations, \eg, complicated backgrounds, obstacle occlusions, distracting objects.

\begin{figure}[t] 
\centering
\hspace*{0.01\linewidth} \\
\vspace{-0.2cm}
\includegraphics[width=0.98\linewidth]{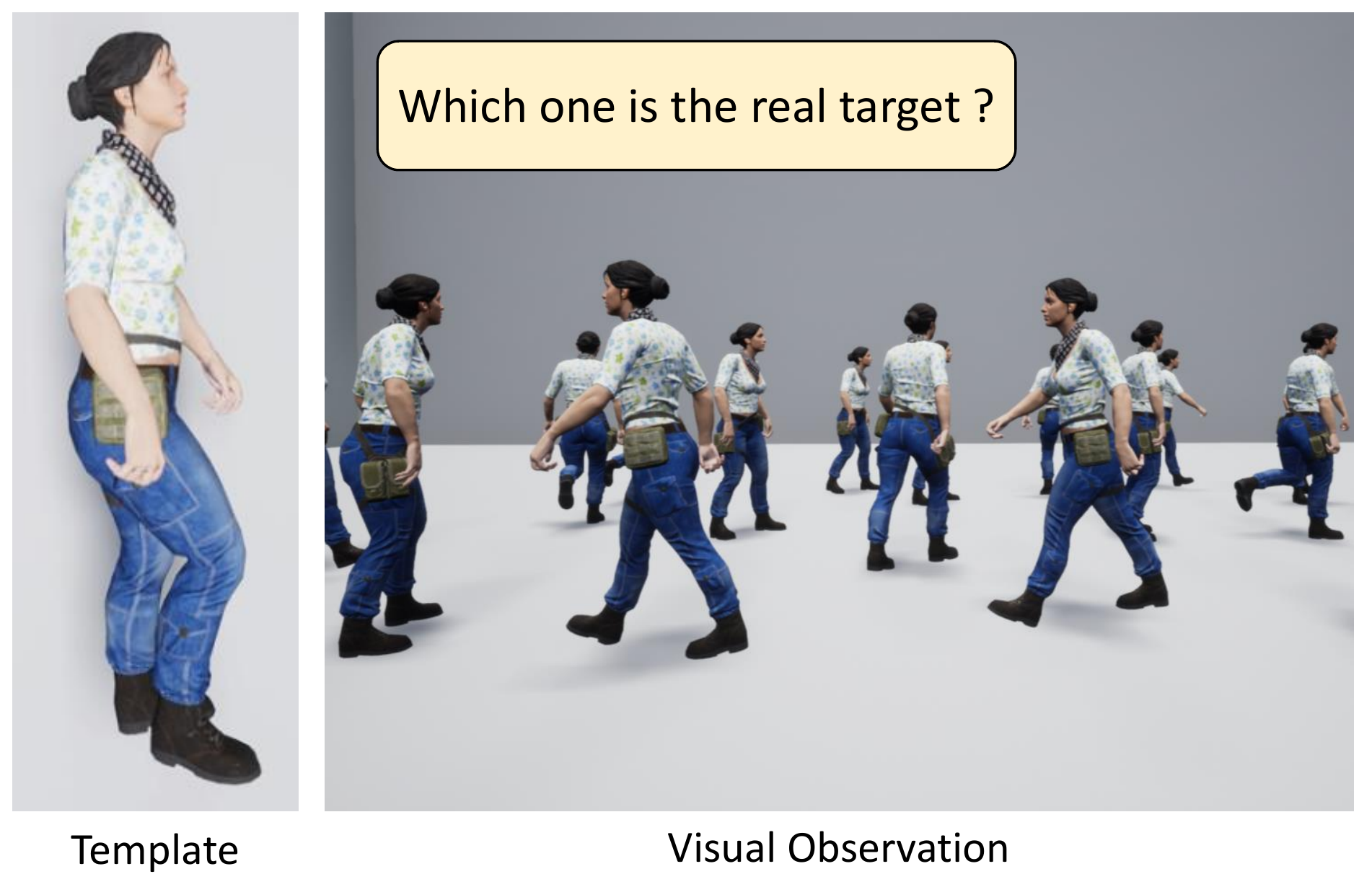}
\vspace{-0.5cm}
\caption{An extreme situation of active visual tracking with distractors. Can you identify which one is the real target?
}
\label{fig:demo}
\vspace{-0.4cm}
\end{figure}

Among all the challenges, the \emph{distractor}, which can induce confusing visual observation and occlusion, is a considerably prominent difficulty. 
Distraction emerges frequently in the real world scenario, such as a school where students wear similar uniforms.
Considering an extreme case (see Fig.~\ref{fig:demo}), there is a group of people dressed in the same clothes, and you are only given a template image of the target, can you confidently identify the target from the crowd?

The distraction has been primarily studied in the \emph{passive} tracking setting~\cite{VOT_TPAMI} (running on collected video clips) but is rarely considered in the \emph{active} setting~\cite{luo2019pami, zhong2019ad}.
In the \emph{passive} setting, previous researchers~\cite{nam2016learning,zhu2018distractor,bhat2019learning} took great efforts on learning a discriminative visual representation in order to identify the target from the crowded background.
Yet it is not sufficient for an \emph{active} visual tracker, which should be of not only a suitable state representation but also an optimal control strategy to move the camera to actively avoid distractions, \eg, finding a more suitable viewpoint to seize the target.

To realize such a robust active tracker, we argue that the primary step is to build an interactive environment, where various distraction situations can frequently emergent.
A straightforward solution is adding a number of moving objects as distractors in the environment.
However, it is non-trivial to model a group of moving objects. 
Pre-defining object trajectories based on hand-crafted rules seems tempting, but it will lead to overfitting in the sense that the yielded tracker generalizes poorly on unseen trajectories.

\begin{figure}[t] 
\centering
\hspace*{0.01\linewidth} \\
\includegraphics[width=0.98\linewidth]{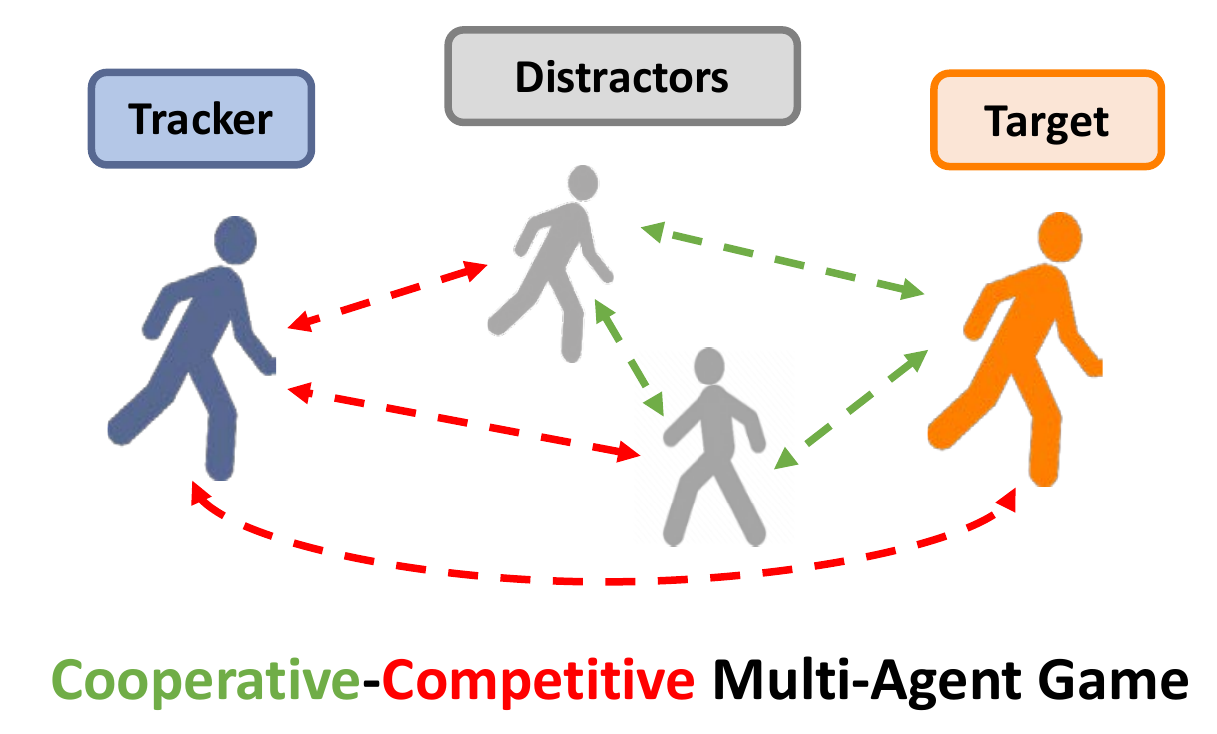}
\vspace{-0.5cm}
\caption{An overview of the Cooperative-Competitive Multi-Agent Game, where the distractors and target cooperate to compete against the tracker.
}
\label{fig:framework}
\vspace{-0.3cm}
\end{figure}

Inspired by the recent work on multi-agent learning~\cite{sukhbaatar2017intrinsic, baker2019emergent}, we propose a mixed Cooperative-Competitive Multi-Agent Game to automatically generate distractions.
In the game, there are three type of agents: tracker, target and distractor. The tracker tries to always follow the target. The target intends to get rid of the tracker. 
The target and distractors form a team, which is to make trouble for the tracker. The cooperative-competitive relations among agents are shown in Fig.~\ref{fig:framework}.
Indeed, it is already challenging to learn an escape policy for a target~\cite{zhong2018advat}.
As the complicated social interaction among agents, it will be even more difficult to model the behavior of multiple objects to produce distraction situations, which is rarely studied in previous work.

Thus, to ease the multi-agent learning process, we introduce a battery of practical alternatives.
First, to mitigate the credit assignments problem among agents, we design a reward structure to explicitly identify the contribution of each distractor with a relative distance factor. 
Second, we propose a cross-modal teacher-student learning strategy, since directly optimizing the visual policies by Reinforcement Learning (RL) is inefficient.
Specifically, we split the training process into two steps.
At the first step, we exploit the grounded state to find meta policies for agents.
Since the grounded state is clean and low-dimensional, we can easily train RL agents to find a equilibrium of the game, \ie, the target actively explores different directions to escape, the distractors frequently appear in the tracker's view, and the tracker still closely follows the target.
Notably, a multi-agent curriculum is also naturally emergent, \ie, the difficulty of the environment induced by the target-distractors cooperation is steadily increased with the evolution of the meta policies.
In the second step, we use the skillful meta tracker (teacher) to supervise the learning of the visual active tracker (student). When the student interacts with the opponents during learning, we replay the multi-agent curriculum by sampling the historical network parameters of the target and distractors. 
Moreover, a recurrent attention mechanism is employed to enhance the state representation of the visual tracker.

The experiments are conducted in virtual environments with numbers of distractors.
We show that our tracker significantly outperforms the state-of-the-art methods in a room with clean backgrounds and a number of moving distractors.
The effectiveness of introduced components are validated in ablation study.
After that, we demonstrate another use of our multi-agent game, adversarial testing the trackers. 
While taking adversarial testing, the target and distractors, optimized by RL, can actively find trajectories to mislead the tracker within a very short time period.
In the end, we validate that the learned policy is of good generalization in unseen environments.
The code and demo videos are available on \url{https://sites.google.com/view/distraction-robust-avt}.

\vspace{-0.3cm}
\section{Related Work}
\vspace{-0.1cm}
\textbf{Active Visual Tracking.}
The methods to realize active visual tracking can be simply divided into two categories: two-stage methods and end-to-end methods. 
Conventionally, the task is accomplished in two cascaded sub-tasks, \ie, the visual object tracking (image $\rightarrow$ bounding box) ~\cite{VOT_TPAMI} and the camera control (bounding box $\rightarrow$ control signal).
Recently, with the advances of deep reinforcement learning and simulation, the end-to-end methods (image $\rightarrow$ control signal), optimizing the neural network in an end-to-end manner~\cite{atari2015}, has achieved a great progress in active object tracking~\cite{luo2018end, li2020pose, zhong2018advat}. It is firstly introduced in ~\cite{luo2018end} to train an end-to-end active tracker in simulator via RL. ~\cite{luo2019end} successfully implements the end-to-end tracker in the real-world scenario. AD-VAT(+)~\cite{zhong2018advat, zhong2019ad} further improves the robustness of the end-to-end tracker with adversarial RL. Following this track, our work is the first paper to study the distractors in AVT with a mixed multi-agent game.

Distractors has attracted attention of researchers in video object tracking~\cite{babenko2009visual,bolme2010visual,kalal2012tracking,nam2016learning,zhu2018distractor}.
To address the issue caused by distractors, data augmentation methods like collecting similar but negative samples to train a more powerful classifier is pretty useful~\cite{babenko2009visual,kalal2012tracking,hare2015struck,zhu2018distractor,deng2020labels}. Other advanced techniques like hard negative examples mining~\cite{nam2016learning} are also proposed for addressing this. Recently, DiMP~\cite{bhat2019learning} develops a discriminative model prediction architecture for tracking, which can be optimized in only a few iterations.
Differently, we model the distractors as agents with learning ability, which can actively produce diverse situations to enhance the learning tracker.

\textbf{Multi-Agent Game.}
It is not a new concept to apply the multi-agent game to build a robust agent.
Roughly speaking, most of previous methods ~\cite{zhong2018advat,held2017automatic,huang2017adversarial,mandlekar2017adversarially,pinto2017robust,sukhbaatar2017intrinsic} focus on modeling a two-agent competition (adversary  protagonist) to learn a more robust policy. The closest one to ours is AD-VAT~\cite{zhong2018advat}, which proposes an asymmetric dueling mechanism (tracker vs. target) for learning a better active tracker.
Our multi-agent game can be viewed as an extension of AD-VAT, by adding a number of learnable distractors within the tracker-target competition.
In this setting, to compete with the tracker, it necessary to learn collaborative strategies among the distractors and target.
However, only a few works~\cite{openai-five,hausknecht2015deep,tampuu2017multiagent,baker2019emergent} explored learning policies under a Mixed Multi-Agent Game.
In these multi-agent games, the agents usually are homogeneous, \ie, each agent plays an equal role in a team.
Differently, in our game, the agents are heterogeneous, including tracker, target, and distractor(s). And the multi-agent competition is also asymmetric, \ie, the active tracker is ``lonely'', which has to independently fight against the team formed by the target and distractors.

  To find the equilibrium in the multi-agent game, usually, Muti-Agent Reinforcement Learning(MARL)~\cite{rashid2018qmix,sunehag2018value,tampuu2017multiagent,lowe2017multi} are employed. Though effective and successful in some toy examples, the training procedure of these methods is really inefficient and unstable, especially in cases that the agents are fed with high-dimensional raw-pixel observation. 
 Recently, researchers have demonstrated that exploiting the grounded state in simulation can greatly improve the stability and speed of vision-based policy training in single-agent scenario~\cite{wilson2020learning,andrychowicz2020learning,pinto2017asymmetric} by constructing a more compact representation or approximating a more precise value function.
  Inspired by these, we exploit the grounded state to facilitate multi-agent learning by taking a cross-modal teacher-student learning, which is close to Multi-Agent Imitation Learning (MAIL).
   MAIL usually rely on a set of collected expert demonstrations~\cite{song2018multi,vsovsic2016inverse,bogert2014multi,lin2014multi} or a programmed expert to provide online demonstrations~\cite{le2017coordinated}.
  However, the demonstrations collection and programmed expert designing are usually performed by human. 
  Instead, we adopt a multi-agent game for better cloning the expert behaviour to a vision-based agent. The expert agent is fed with the grounded state and learned by self-play.

\vspace{-0.2cm}
\section{Multi-Agent Game}
Inspired by the AD-VAT~\cite{zhong2018advat}, we introduce a group of active distractors in the tracker-target competition to induce distractions. 
We model such a competition as a mixed Cooperative-Competitive Multi-Agent Game, where agents are employed to represent the tracker, target and distractor, respectively.
In this game, the target and distractor(s) constitute a cooperative group to actively find the weakness of the tracker. 
In contrast, the tracker has to compete against the cooperative group to continuously follow the target. 
To be specific, each one has its own goal, shown as following:
\begin{itemize}
\vspace{-0.3cm}
    \item \emph{Tracker} chases the target object and keeps a specific relative distance and angle from it.
        \vspace{-0.1cm}
    \item \emph{Target} finds a way to get rid of the tracker.
    \vspace{-0.1cm}
    \item \emph{Distractor} cooperates with the target and other distractors to help the target escape from the tracker, by inducing confusing visual observation or occlusion.
\end{itemize}
\vspace{-0.3cm}
\subsection{Formulation}
Formally, we adopt the settings of Multi-Agent Markov Game~\cite{littman1994markov}.
Let subscript $i \in \{1, 2, 3\}$ be the index of each agent,
\ie, $i=1$, $i=2$, and $i=3$ denote the tracker, the target, and the distractors, respectively.
Note that, even the number of distractors would be more than one, we use only one agent to represent them. That is because they are homogeneous and share the same policy. 
The game is governed by the tuple $<\FS, \FO_i, \FA_i,  \FR_i, \FP>$,
$i=1,2,3$,
where $\FS, \FO_i, \FA_i, \FR_i, \FP$ denote the joint state space, the observation space (agent $i$), the action space (agent $i$), the reward function (agent $i$) and the environment state transition probability, respectively.
Let a secondary subscript $t \in \{1, 2,...\}$ denote the time step.
In the case of visual observation, we have each agent's observation $o_{i, t} = o_{i, t}(s_t, s_{t-1}, o_{i, t-1})$, where $o_{i, t}, o_{i, t-1} \in \FO_{i}$, $s_t, s_{t-1} \in \FS$. 
Since the visual observation is imperfect, the agents play a partially observable multi-agent game.
It reduces to $o_{i,t} = s_t$ in the case of fully observable game, which means that the agent can access the grounded states directly.
In the AVT task, the grounded states are the relative poses among all agents, needed by the meta policies.
When all the three agents take simultaneous actions $a_{i,t} \in \FA_{i}$,
the updated state $s_{t+1}$ is drawn from the environment state transition probability, as $s_{t+1} \sim \FP(\cdot|s_t, a_{1,t}, a_{2,t}, a_{3,t})$.
Meanwhile, each agent receives an immediate reward $r_{i,t} = r_{i,t}(s_t, a_{i,t})$ respectively.
The policy of the agent $i$, $\pi_{i}(a_{i,t}|o_{i,t})$, is a distribution over its action $a_{i,t}$ conditioned on its observation $o_{i,t}$.
Each policy $\pi_i$ is to maximize its cumulative discounted reward $\mathbf{E}_{\pi_i} \left[ \sum_{t=1}^{T} \gamma^{t-1} r_{i,t} \right]$, where $T$ denotes the horizontal length of an episode and $r_{i,t}$ is the immediate reward of agent $i$ at time step $t$.
The policy takes as function approximator a neural network with parameter $\Theta_{i}$, written as $\pi_{i}(a_{i,t}|o_{i,t};\Theta_{i})$.
The cooperation-competition will manifest by the design of reward function $r_{i,t}$, as described in the next subsection.

\subsection{Reward Structure}
\label{sec:reward}

\begin{figure}[t]
\centering
\hspace*{0.01\linewidth} \\
\includegraphics[width=0.8\linewidth]{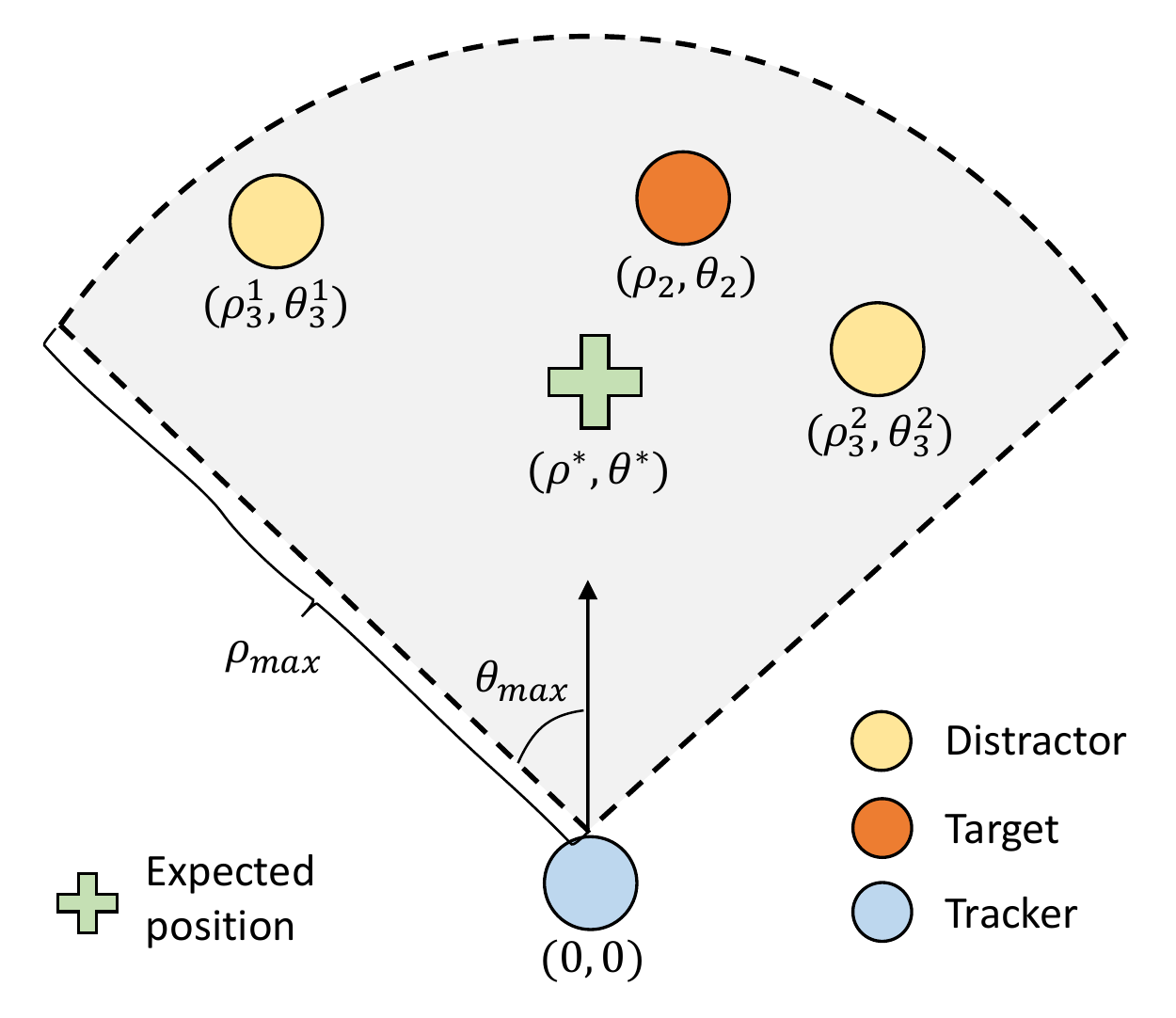}
\vspace{-0.4cm}
\caption{A top-down view of the tracker-centric coordinate system with target (orange), distractors (yellow), and expected target position (green). Tracker (blue) is at the origin of coordinate $(0, 0)$. The gray sector area represents the observable area of the tracker. The arrow in the tracker notes the front of the camera.
}
\vspace{-0.4cm}
\label{fig:reward}

\end{figure}

To avoid the aforementioned intra-team ambiguity, we propose a distraction-aware reward structure, which modifies the cooperative-competitive reward principle to take into account the following intuition: 
1) The target and distractors are of a common goal (make the tracker fail). So they need to share the target's reward to encourage the target-distractor cooperation. 
2) Each distractor should have its own rewards, which can measure its contribution to the team.
3) The distractor obviously observed by the tracker will cause distraction in all probability, and the one that out of the tracker's  view can by no means ``mislead" the tracker.

We begin with defining a tracker-centric relative distance $d(1, i)$, measuring the geometric relation between the tracker and the other player $i>1$.
\begin{equation}
\label{eq:tracker_reward}
	d(1,i) = \frac{\vert\rho_i - \rho^*\vert}{\rho_{max}} + \frac{\vert\theta_{i}-\theta^*\vert}{\theta_{max}},
\end{equation}
where $(\rho_i, \theta_i)$ and $(\rho^*, \theta^*)$ represent the location of the player $i > 1$ and the expected target in a tracker-centric polar coordinate system, respectively. $\rho$ is the distance from the origin (tracker), and $\theta$ is the relative angle to the front of the tracker. 
See Fig.~\ref{fig:reward} for an illustration.

With the relative distance, we now give a formal definition of the reward structure as:
\begin{equation}
\begin{array}{ll}
r_1 = 1 - d(1, 2)\,,\\
r_2 = - r_1\,,\\
r_{3}^j =  r_2 - d(1, j)\,.\\
\end{array}
\end{equation}
Here we omit the timestep subscript $t$ without confusion.
The tracker reward is similar to AD-VAT~\cite{zhong2018advat}, measured by the distance between the target and expected location. The tracker and target play a zero-sum game, where $r_1 + r_2 = 0$. The distractor is to cooperate with the target by sharing $r_2$. Meanwhile, we identify its unique contribution by taking its relative distance $d(1,j)$ as a penalty term in the reward. It is based on an observation that once the tracker is misled by a distractor, the distractor will be regarded as the target and placed at the center of the tracker's view. Otherwise, the distractor will be penalized by $d(i, j)$ when it is far from the tracker, as its contribution to the gain of $r_2$ will be marginal. Intuitively, the penalty term $d(1,j)$ can guide the distractors learn to navigate to the tracker's view, and the bonus from target $r_2$ can encourage it to cooperate with the target to produce distraction situations to mislead the tracker.
In the view of relative distance, the distractor is to minimize $\sum_{t=1}^{T} d_t(1, j)$ and maximize $\sum_{t=1}^{T} d_t(1, 2)$, while the tracker is to minimize $\sum_{t=1}^{T} d_t(1, 2)$.
Besides, if a collision is detected to agent $i$, we penalize the agent with a reward of $-1$. When we remove the penalty, the learned distractors would prefer to physically surround and block the tracker , rather than make confusing visual observations to mislead the tracker.

\section{Learning Strategy}
To efficiently learn policies in the multi-agent game, we introduce a two-step learning strategy to combine the advantages of Reinforcement Learning (RL) and Imitation Learning (IL).
First, we train meta policies (using the grounded state as input) via RL in a self-play manner.
After that, IL is employed to efficiently impart the knowledge learned by meta policies to the active visual tracker.
Using the grounded state can easily find optimal policies (teacher) for each agent first. The teacher can guide the learning of the visual tracker (student) to avoid numerous trial-and-error explorations. 
Meanwhile, opponent policies emergent in different learning stage are of different level of difficulties, forming a curriculum for the student learning.

\subsection{Learning Meta Policies with Grounded State}
\label{learng-teacher}

At the first step, we train meta policies using self-play.
Hence, the agent can always play with opponents of an appropriate level, regarded as a natural curriculum~\cite{baker2019emergent}. 
The \emph{meta policies}, noted as $\pi_1^*(s_t)$, $\pi_2^*(s_t)$, $\pi_3^*(s_t)$, enjoy privileges to access the grounded state $s_t$, rather than only visual observations $o_{i,t}$.
Even though such grounded states are unavailable in most real-world scenarios, we can easily reach it in the virtual environment.
For AVT, the grounded state is the relative poses (position and orientation) among players.
We omit the shape and size of the target and distractors, as they are similar during training.
Note that the state for agent $i$ will be transformed into the entity-centric coordinate system before feed into the policy network.
To be specific, the input of agent $i$ is a sequence about the relative poses to other agents, represented as ${P_{i,1}, P_{i,2}, ... P_{i,n}}$, where $n$ is the number of the agents and $P_{i.j}=(\rho_{i,j}, cos(\theta_{i,j}), sin(\theta_{i,j}), cos(\phi_{i,j}), sin(\phi_{i,j})$. Note that $(\rho_{i,j}, \theta_{i,j}, \phi_{i,j})$ indicates the relative distance, angle, and relative orientation from agent $i$ to agent $j$. Agent $i$ is at the origin of the coordination.
Since the number of the distractors is randomized during either training or testing, the length of the input sequence would be different across each episode. Thus, we adopt the Bidirectional-Gated Recurrent Unit (Bi-GRU) to pin-down a fixed-length feature vector, to enable the network to handle the variable-length distractors. Inspired by the tracker-award model in AD-VAT~\cite{zhong2018advat}, we also fed the tracker's action $a_1$ to the target and distractors, to induce stronger adversaries.

During training, we optimize the meta policies with a modern RL algorithm, \eg, A3C~\cite{mnih2016asynchronous}.
To collect a model pool containing policies at different levels, we save the network parameters the of target and distractor every $50K$ interactions.
During the student learning stage, we can sample the old parameters from the model pool for the target and distractors to reproduce the emergent multi-agent curriculum.
Note that we further fine-tune the meta trackers to play against all the opponents in the model pool before going into the next stage.
More details about the meta policies can be found in Appendix. A.

\subsection{Learning Active Visual Tracking}
\label{Sec:AL}

\begin{figure}[t]
\centering
\hspace*{0.01\linewidth} \\
\includegraphics[width=0.98\linewidth]{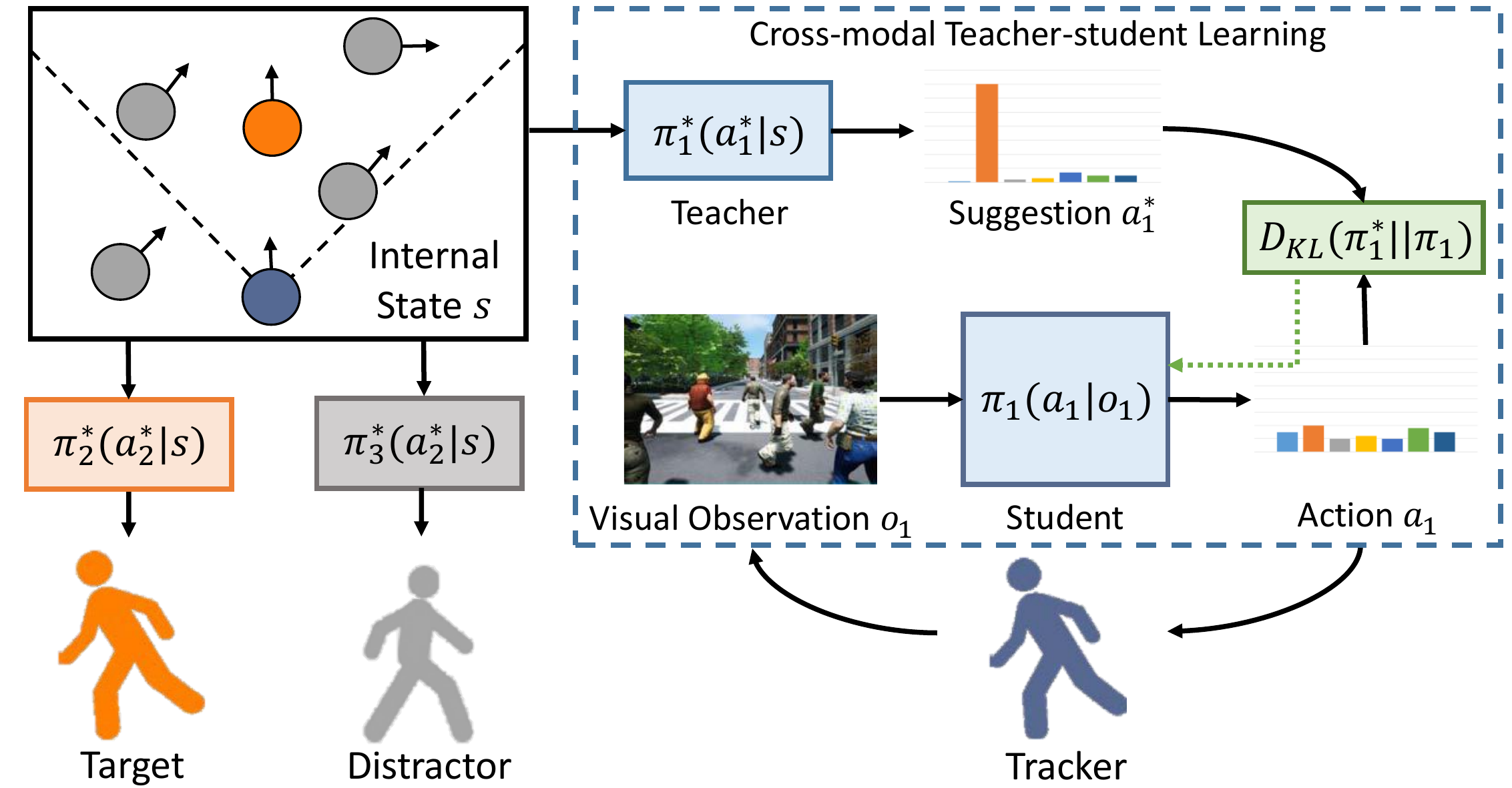}
\caption{An overview of the cross-modal teacher-student learning strategy. Blue, orange, gray represent tracker, target, distractors, respectively. $\pi_1^*$, $\pi_2^*$, $\pi_3^*$ enjoy privileges to acquire the grounded state as input. 
The tracker adopts the student network (visual tracker) to plays against opponents (target and distractors) to collect useful experiences for learning.
We sample parameters from the model pool constructed in the first stage.
During training, the student network is optimized by minimizing the KL divergence between the teacher suggestion and the student output.
}
\label{fig:learning}
\end{figure}

With the learned meta policies, we seek out a teacher-student learning strategy to efficiently build a distraction-robust active visual tracker in an end-to-end manner, shown as Fig.~\ref{fig:learning}. 
We apply the meta tracker $\pi_1^*(s_t)$ (teacher) to teach a visual tracker $\pi_1(o_{i,t})$ (student) to track. In the teacher-student learning paradigm, the student needs to clone the teacher's behavior. 
Therefore, we dive into the behavioral cloning problem.
However, it is infeasible to directly apply supervised learning to learn from the demonstration collected by expert's behavior. 
Because the learned policy will inevitably make at least occasional mistakes.
However, such a small error may lead the agent to a state which deviates from expert demonstrations.
Consequently, The agent will make further mistakes, leading to poor performance.
At the end, the student will be of poor generalization to novel scenes. 

Thus, we take an interactive training manner as DAGGER~\cite{ross2011reduction}, in which the student takes actions from the learning policy and gets suggestions from the teacher to optimize the policy.
To be specific, the training mechanism is composed of two key modules: \emph{Sampler} and \emph{Learner}. 
In the \emph{Sampler}, we perform the learning policy $\pi_1(o_{i,t})$ to control the tracker to interact with the others.
The target and distractors are governed by meta policies $\pi_2^*(s_t)$ and $\pi_3^*(s_t)$ respectively.
Meanwhile, the meta tracker $\pi_1^*(s_t)$ provides expert suggestions $a_{1,t}^*$ by monitoring the grounded state. 
At each step, we sequentially store the visual observation and the expert suggestions $(o_{1,t}, a_{1,t}^*)$ in a buffer $B$. 
To make diverse multi-agent environments, we random sample parameters from the model pool for $\pi_2^*(s_t)$ and $\pi_3^*(s_t)$.
The model pool is constructed during the first stage, containing meta policies at different levels.
So we easily reproduce the multi-agent curriculum emergent in the first stage.
We also demonstrate the importance of the multi-agent curriculum in the ablation analysis.

In parallel, the \emph{Learner} samples a batch of sequences from the buffer $B$ and optimizes the student network in a supervised learning manner.
The objective function of the learner is to minimize the relative entropy (Kullback-Leibler divergence) of the action distribution between student and teacher, computed as:
\begin{equation}
    \mathcal{L}_{KL} = \sum_{n=1}^{N} \sum_{t=1}^{T} D_{KL}(a_{1,t}^*||\pi(o_{t}))\,,
    \label{kl}
\end{equation}
where $N$ is the number of trajectories in the sampled batch, $T$ is the length of one trajectory.
In practice, multiple samplers and one learner work asynchronously, significantly reducing the time needed to obtain satisfactory performance.

Moreover, we employ a recurrent attention mechanism in the end-to-end tracker network~\cite{ zhong2018advat} to learn a representation which is consistent in spatial and temporal.
We argue that a spatial-temporal representation is needed for the active visual tracking, especially in the case of distraction appearing.
Specially, we use the ConvLSTM~\cite{xingjian2015convolutional} to encode an attention map, which is multiplied by the feature map extracted a target-aware feature from the CNN encoder. See Appendix.B for more details.

\section{Experimental Setup}
In this section, we introduce the environments, baselines, and evaluation metrics used in our experiments.

\begin{figure}[t]
\centering
\hspace*{0.01\linewidth} \\
\includegraphics[width=0.48\linewidth]{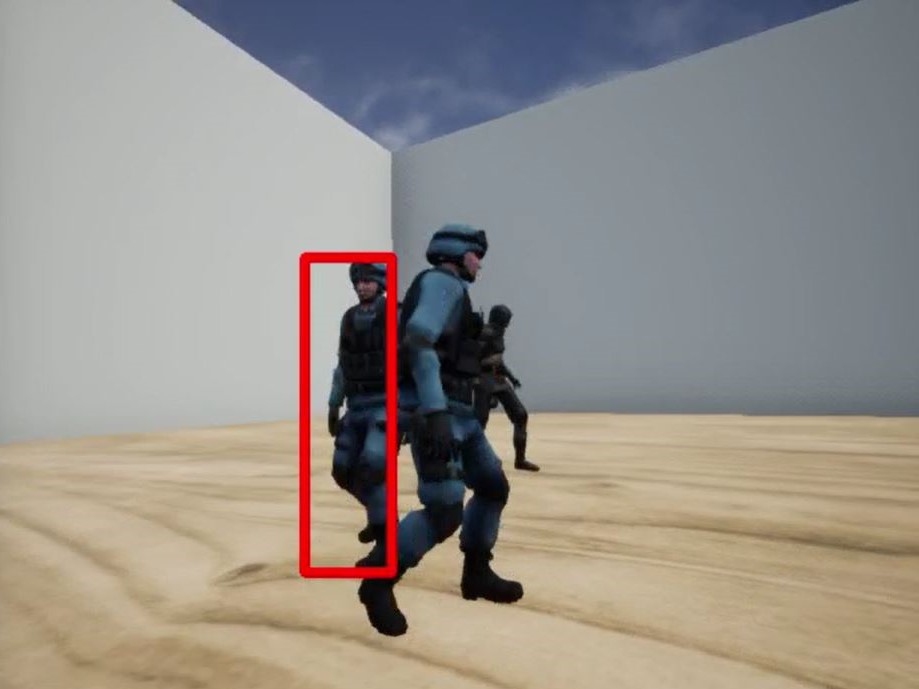}
\includegraphics[width=0.48\linewidth]{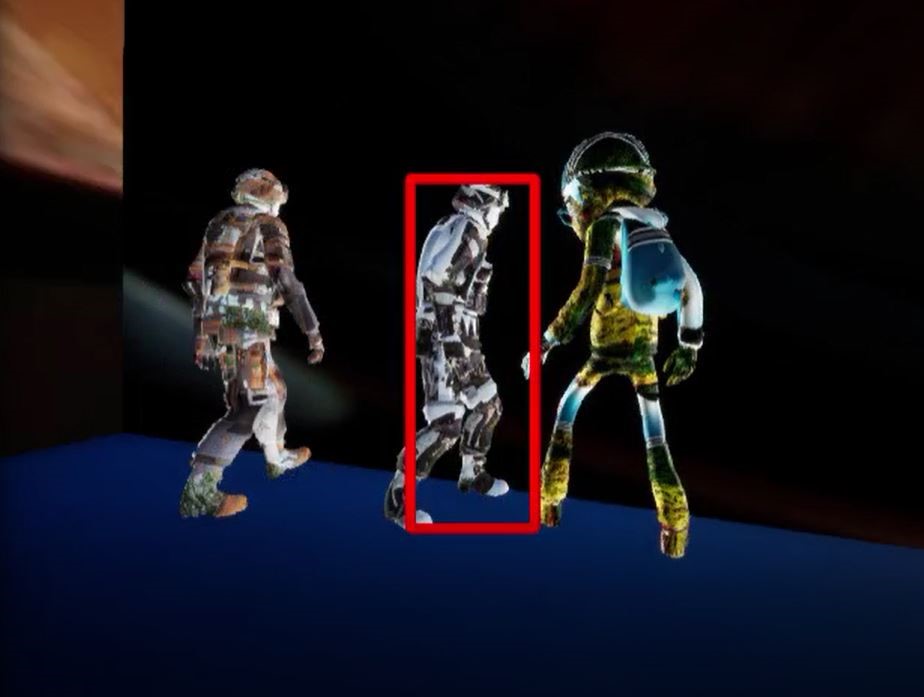}
\vspace{-0.2cm}
\caption{The snapshots of tracker's visual observation in ~\emph{Simple Room}. The right is the augmented training environment. The target is pointed out with a bounding box.
}
\label{fig:simple_room}
\vspace{-0.3cm}
\end{figure}

\textbf{Environments.}
Similar to previous works~\cite{luo2019pami,zhong2018advat}, the experiments are conducted on  UnrealCV environments~\cite{qiu2017unrealcv}.
We extend the two-agent environments used in AD-VAT~\cite{zhong2018advat} to study the multi-agent ($n>2$) game. 
Similar to AD-VAT, the action space is discrete with seven actions, \emph{move-forward/backward, turn-left/right, move-forward-and-turn-left/right,} and \emph{no-op}. 
The observation for the visual tracker is the color image in its first-person view.
The primary difference is that we add a number of controllable distractors in the environment, shown as Fig.~\ref{fig:simple_room}.
Both target and distractors are controlled by the scripted navigator, which temporally sample a free space in the map and navigate to it with a random set velocity.
Note that we name the environments that use the scripted navigator to control the target and $x$ distractors as \emph{Nav-x}. If they are governed by the meta policies, we mark the environment as \emph{Meta-x}.
Besides, we enable agents to access the poses of players, which is needed by the meta policies.
Two realistic scenarios (\emph{Urban City} and \emph{Parking Lot}) are used to verify the generalization of our tracker in other unseen realistic environments with considerable complexity.
In \emph{Urban City}, there are five unseen characters are placed, and the appearance of each is randomly sampled from four candidates. So it is potential to see that two characters dressed the same in an environment.
In \emph{Parking Lot}, all of the target and distractors are of the same appearance.
Under this setting, it would be difficult for the tracker to identify the target from distractions.

\textbf{Evaluation Metrics.}
We employ the metrics of Accumulated Reward (AR), Episode Length (EL), Success Rate (SR) for our evaluation.
Among those metrics, AR is the recommended primary
measure of tracking performance, as it considers both
precision and robustness. The other metrics are also reported
as auxiliary measures.
Specifically,
AR is affected by the immediate reward and the episode length.
Immediate reward measures the goodness of tracking at each step. 
EL roughly measures the duration of good tracking, as the episode is terminated when the target is lost for continuous $5$ seconds or it reaches the max episode length.
SR is employed in this work to better evaluate the robustness, which counts the rate of successful tracking episodes after running $100$ testing episodes.
An episode is marked as success only if the tracker continuously follows the target till the end of the episode (reaching the max episode length).

\textbf{Baselines.}
We compare our method with a number of state-of-the-art methods and their variants, including the two-stage and end-to-end trackers.
First, We develop conventional two-stage active tracking methods by combining passive trackers with a PID-like controller.
As for the passive trackers, we directly use three off-the-shelf models (DaSiamRPN~\cite{zhu2018distractor}, ATOM~\cite{danelljan2019atom}, DiMP~\cite{bhat2019learning}) without additional training in our environment.
Notably, both DiMP and ATOM can be optimized on the fly to adapt to a novel domain.
So they can generalize well to our virtual environments and achieve strong performance in the no-distractor environments, \eg, DiMP tracker achieves $1.0$ SR in \emph{Simple Room (Nav-0)}.
Second, two recent end-to-end methods (SARL~\cite{luo2019pami}, AD-VAT~\cite{zhong2018advat}) are reproduced in our environment to compare. We also extend them by adding two random walking distractors in the training environment, noted as SARL+ and AD-VAT+.

All end-to-end methods are trained in \emph{Simple Room} with environment augmentation~\cite{luo2019pami}.
After the learning converged, we choose the model that achieves the best performance in the environment for further evaluation.
Considering the random factors, we report the average results after running $100$ episodes. 
More implementation details are introduced in Appendix.C.

\vspace{-0.2cm}
\section{Results}
\vspace{-0.1cm}
We first demonstrate the evolution of the meta policies while learning in our game. Then, we report the testing results in \emph{Simple Room} with different numbers of distractors.
After that, we conduct an ablation study to verify the contribution of each component in our method.
We also adversarially test the trackers in our game.
Moreover, we evaluate the transferability of the tracker in photo-realistic environments.

\subsection{The Evolution of the Meta Policies}
\begin{figure}[t]
\centering
\hspace*{0.01\linewidth} \\
\includegraphics[width=0.98\linewidth]{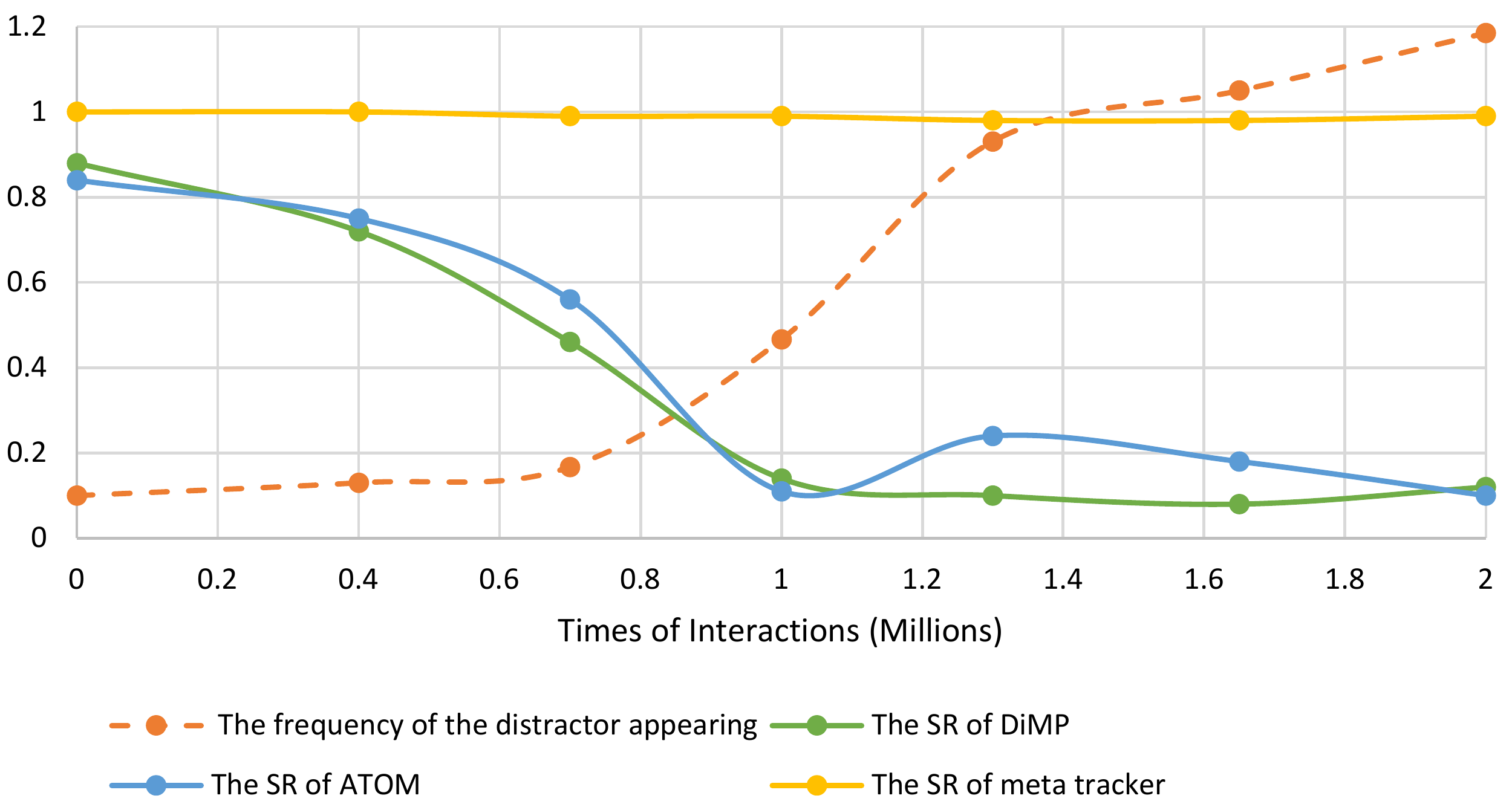}
\caption{The evolution of target-distractor cooperation in the multi-agent game.}
\label{fig:evolution}
\vspace{-0.5cm}
\end{figure}
While learning to play the multi-agent game, the multi-agent curriculum automatically emerges.
To demonstrate it, we evaluate the skill-level of the adversaries (target+distractors) at different learning stages from two aspects: the frequency of the distractor appearing in the tracker's view and the success rate (SR) of the off-the-shelf trackers (DiMP and ATOM). 
To do it, we collects seven meta policies after agents take $0$, $0.4M$, $0.7M$, $1M$, $1.3M$, $1.65M$, $2M$ interactions, respectively. 
We then make the visual tracker (ATOM and DiMP) to play with each collected adversaries (one target and two distractors) in \emph{Simple Room}, and count the success rate of each tracker in 100 episodes. 
we also let the converged meta tracker (at $2M$) follow the target-distractors group from different stages, respectively. 
And we report the success rate of the meta tracker and the frequency of the distractor appearing in the tracker's view, shown as Fig.~\ref{fig:evolution}.
We can see that the frequency of the distractor appearing is increased during the multi-agent learning. Meanwhile, the success rate of the visual tracker is decreased. 
This evidence shows that the complexity of the multi-agent environment is steadily increased with the development of the target-distractor cooperation.
Meanwhile, we also notice that the learned meta tracker can robustly follow the target ($SR \ge 0.98$), even when the distractors frequently appearing in its view.
This motivates us to take the meta tracker as a teacher to guide the learning of the active visual tracker.

\subsection{Evaluating with Scripted Distractors}
\label{compared with baseline}

\begin{figure}[tb]
\centering
\hspace*{0.01\linewidth} \\
\includegraphics[width=0.98\linewidth]{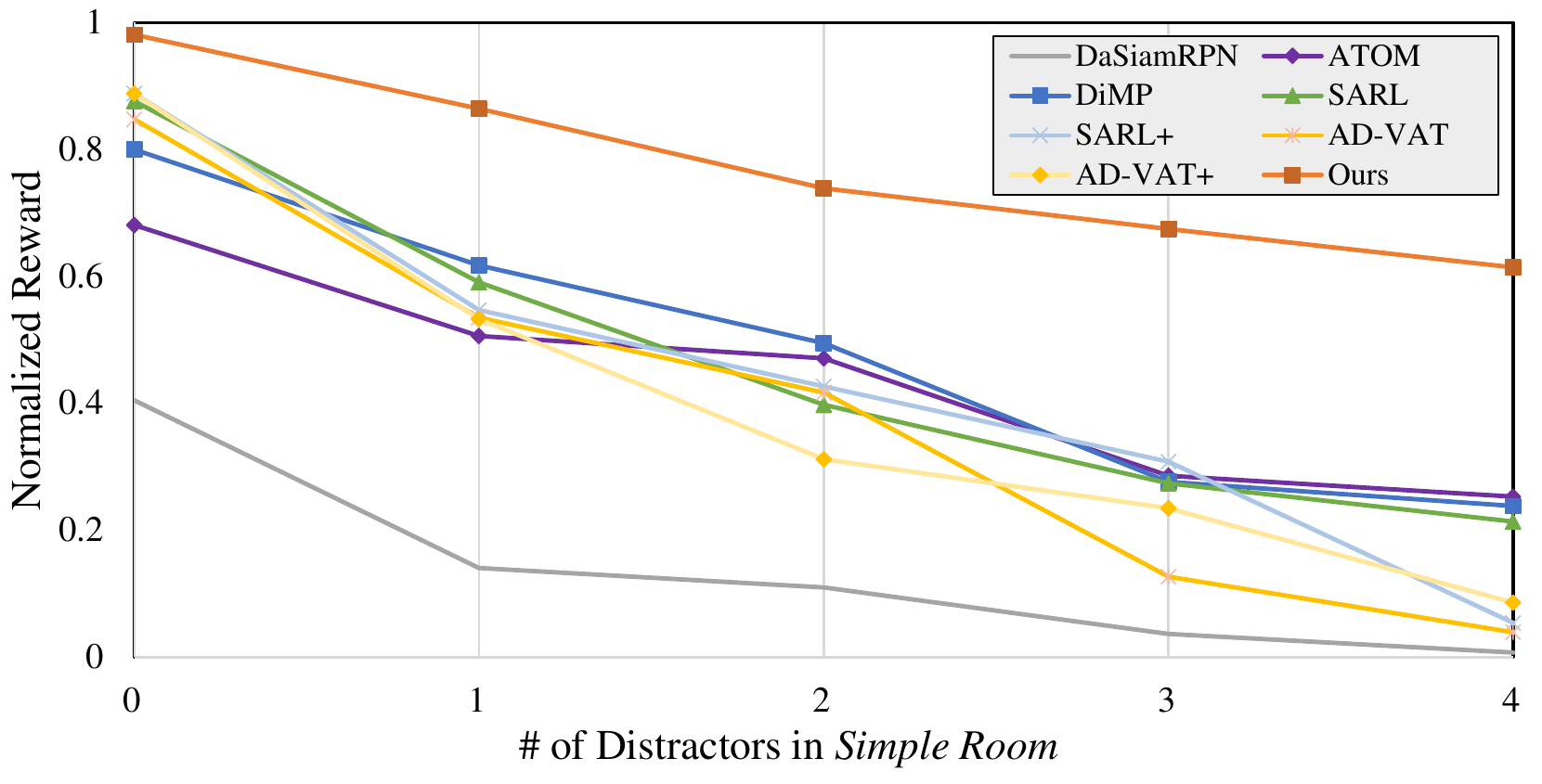}
\caption{ Evaluating the distraction-robustness of the trackers by increasing the number of random distractors in \emph{Simple Room}}
\vspace{-0.3cm}
\label{fig:simpleroom}
\end{figure}

We analyze the distraction robustness of the tracker in \emph{Simple Room} with scripted target and distractors. 
This environment is relatively simple to most real-world scenarios as the background is plain and no obstacles is placed.
So most trackers can precisely follow the target when there is no presence of distraction.
Hence, we can explicitly analyze the distraction robustness by observing how the tracker's performance is changed with the increasing number of distractors, shown as Fig.~\ref{fig:simpleroom}.
Note that we normalize the reward by the average score achieved by the meta tracker, which is regarded as the best performance that the tracker could reach in each configuration.
We observed that the learned meta tracker is strong enough to handle different cases, and hardly lost in the environment.

We can see that most methods (except DaSiamRPN) are competitive when there is no distractor. 
The low score of DaSiamRPN is mainly due to the inaccuracy of the predicted bounding box, which further leads to the tracker's failure in keeping a certain distance from the target. 
With the increasing number of distractors, the gap between our tracker and baselines gradually broaden.
For example, in the four distractors room, the normalized reward achieved by our tracker is two times the ATOM tracker, \ie, $0.61$ vs. $0.25$.
For the two-stage methods, in most simple cases, DiMP is a little better than ATOM, thanks to the discriminative model prediction. 
However, it remains to get lost easily when the distractor occludes the target.
We argue that there are two reasons leading to the poor performance of the passive tracker when distractions appearing:
1) the target representation is without long-term temporal context.
2) the tracker lacks a mechanism to predict the state of an invisible target.
Thus, once the target is disappeared (occluded by distractors or out of the view), the tracker will regard the observed distractor as a target.
If it follows the false target to go, the true target will hardly appear in its view again. 

For the end-to-end methods, it seems much weaker than the two-stage methods in the distraction robustness, especially when playing against many distractors.
By visualizing the tracking process, we find that AD-VAT tends to follow the moving object in the view but unable to identify which is the true target.
So it is frequently misled by the moving objects around the target.
Besides, the curves of SARL+ and AD-VAT+ are very close to the original version (SARL and AD-VAT). 
However, without the target-distractor cooperation, the distraction situation appears at a low frequency in the plus versions. 
Thus, the learned trackers are still vulnerable to the distractors, and the improvements they achieved are marginal.
This indicates that it is useless to simply augment the training environment with random moving distractors.

Our tracker significantly outperforms others in all the cases of distractors.
This evidence shows that our proposed method is of great potential for realizing robust active visual tracking in a complex environment.
However, the performance gap between our model and the teacher model ($1 - ours$) indicates that there is still room for improvement. By visualizing the test sequences of our model we find that it mainly fails in extreme tough cases where the tracker is surrounded by some distractors that totally occlude the target or block the way to track. More vivid examples can be found in the demo video.

\subsection{Ablation Study}
\label{abalation}

\begin{table}[t]
\caption{Ablative analysis of the visual policy learning method in \emph{Simple Room}. The best results are shown in bold.}
\resizebox{\linewidth}{!}{
\begin{tabular}{cll|ccc|ccc}
\hline
\multicolumn{3}{c|}{Methods}     & \multicolumn{3}{c|}{Nav-4} & \multicolumn{3}{c}{Meta-2} \\
\multicolumn{3}{c|}{}  & AR      & EL      & SR    & AR     & EL       & SR     \\
\hline
\multicolumn{3}{c|}{Ours}                         & \bf250   & \bf401   & \bf0.54  & \bf141    & \bf396    & \bf0.44 \\
\hline
\multicolumn{3}{c|}{w/o multi-agent curriculum}   & 232   & 394     & 0.53  & -23    & 283   & 0.08   \\
\multicolumn{3}{c|}{w/o teacher-student learning} & 76      & 290     & 0.22  & 79     & 340    & 0.4    \\
\multicolumn{3}{c|}{w/o recurrent attention}      & 128     & 193     & 0.27  & 75     & 331      & 0.32  \\
\hline
\end{tabular}
}
\end{table}
 We conduct an ablation study to better understand the contribution of each introduced component in our learning method.
1) To evaluate the effectiveness of the multi-agent curriculum, we use the scripted navigator to control the target and distractors when taking the teacher-student learning, instead of replaying the policies collected in the model pool.
We find that the learned tracker obtains comparable results to ours in the Nav-4, but there is an obvious gap in Meta-2, where the target and 2 distractors are controlled by the adversarial meta policies. 
This shows that the tracker over-fits specific moving pattern of the target, but does not learn the essence of active visual tracking.
2) For teacher-student learning, we directly optimize the visual tracking network by A3C, instead of using the suggestions from the meta tracker.
Notably, such method can also be regarded as a method that augments the environment by multi-agent curriculum for SARL method. So, by comparing its result with SARL, we can also recognize the value of the multi-agent curriculum on improving the distraction robustness.
For the recurrent attention, we compare it with the previous Conv-LSTM network introduced in \cite{zhong2018advat}. We can see that the recurrent attention mechanism can significantl improve the performance of the tracker in both settings.

\vspace{-0.1cm}
\subsection{Adversarial Testing}
Beyond training a tracker, our multi-agent game can also be used as a test bed to further benchmark the distraction robustness of the active trackers. 
In adversarial testing, the target collaborates with distractors to actively find adversarial trajectories that fail the tracker to follow. 
Such an adversarial testing is necessary for AVT.
Because the trajectories generated by rule-based moving objects designed for evaluation can never cover all of the possible cases, \ie the trajectories of objects can be arbitrary and have infinitely possible patterns. 
Moreover, it can also help us discover and understand the weakness of the learned tracker, thus facilitating further development.

\begin{figure}[tb]
\centering
\hspace*{0.01\linewidth} \\
\includegraphics[width=\linewidth]{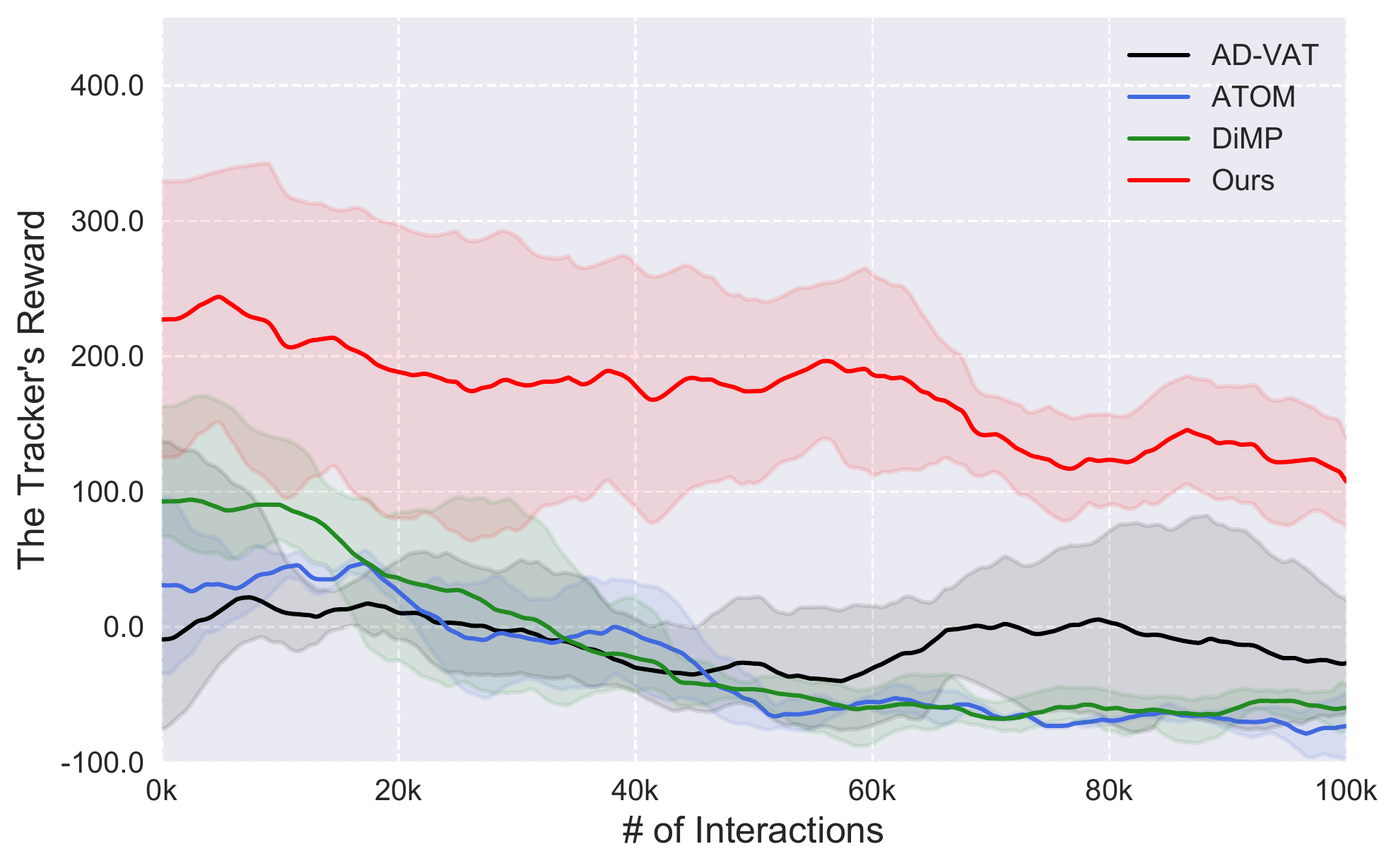}
\vspace{-0.7cm}
\caption{
The reward curves of four trackers during the adversarial testing, running with three random seeds. Better viewed in color.
}
\label{fig:adv_test}
\vspace{-0.3cm}
\end{figure}

We conduct the adversarial testing by training the adversaries to find model-specific adversarial trajectories for each tracker.
In this stage, the network of target and distractors are initialized with parameters from the meta polices. 
For a fair comparison, We iterate the adversaries in $100K$ interaction samples. 
The model of the tracker is frozen during the adversarial testing.
The adversarial testing is conduct in ~\emph{Simple Room} with $2$ distractors.
The curves are the average of three runs with different random seeds, and the shaded areas are the standard errors of the mean.
Fig.~\ref{fig:adv_test} plots the reward of four trackers during the adversarial testing. 

We find that most of the trackers are vulnerable to adversaries, resulting in a fast descending of the model's reward.
The rewards of all the methods drop during the testing, showing that the adversaries are learning a more competitive behaviour.
The reward of our tracker leads the baseline methods most of the time.
In the end, DiMP and ATOM are struggling in the adversarial case, getting very low rewards, ranging from $-60$ to $-100$.

Besides, we also observe an interesting but difficult case. The target rotates at a location and the distractors move around the target and occlude the target; After a while, the distractor goes away, and the two-stage trackers will follow the distractor instead of the target. The demo sequences are available in the demo video.
The adversarial testing provides a new evidence to the robustness of our tracker. It also reflects the effectiveness of our method in learning target-distractor collaboration.

\subsection{Transferring to Unseen Environments}

To show the potential of our model in realistic scenarios, we validate the transferability of the learned model in two photo-realistic environments, which are distinct from the training environment.

\begin{figure}[tb]
\centering
\hspace*{0.01\linewidth} \\
\includegraphics[width=\linewidth]{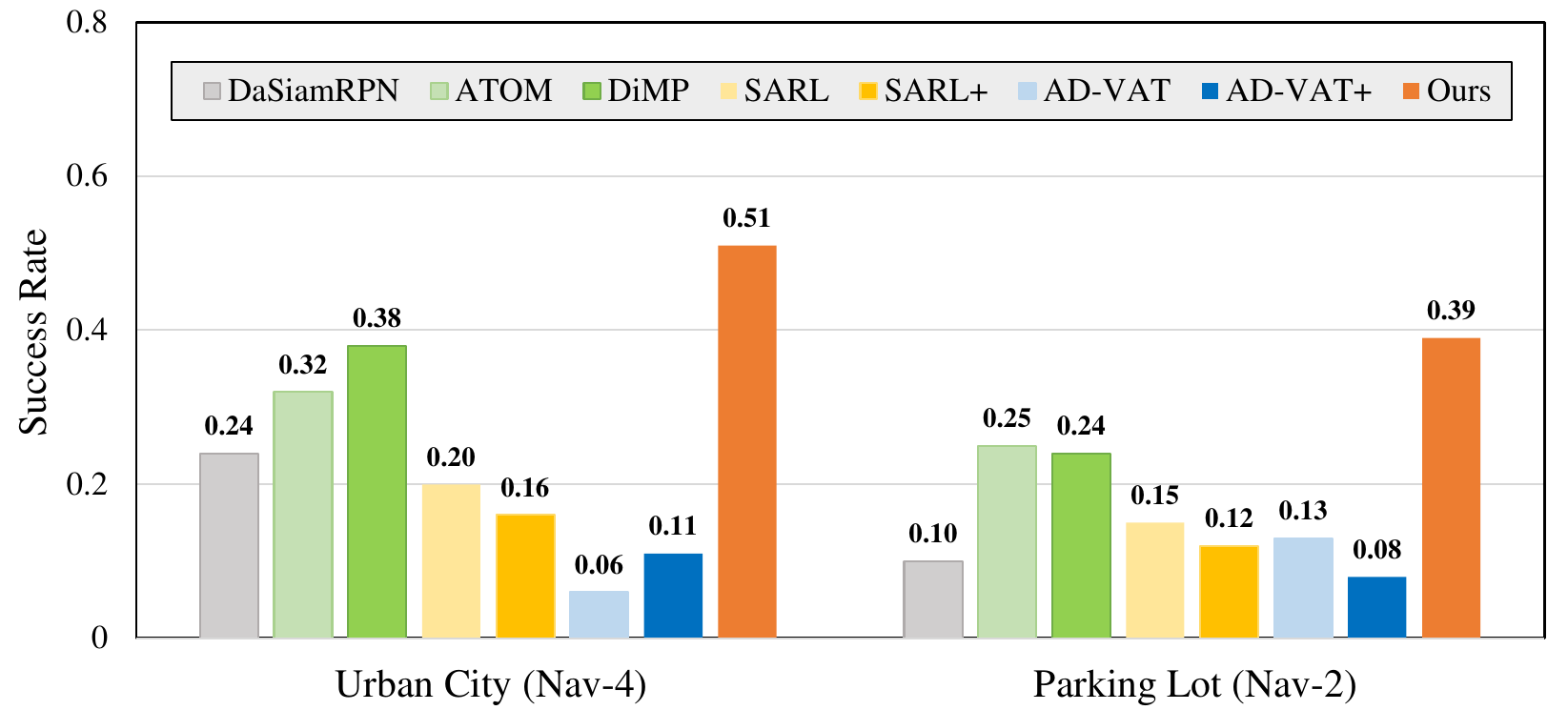}
\vspace{-0.4cm}
\caption{Evaluating generalization of the tracker on two unseen environments (\emph{Urban City} and \emph{Parking Lot}).
}
\label{fig:generalization}
\vspace{-0.3cm}
\end{figure}

\begin{figure}[tb]
\centering
\hspace*{0.01\linewidth} \\
\includegraphics[width=\linewidth]{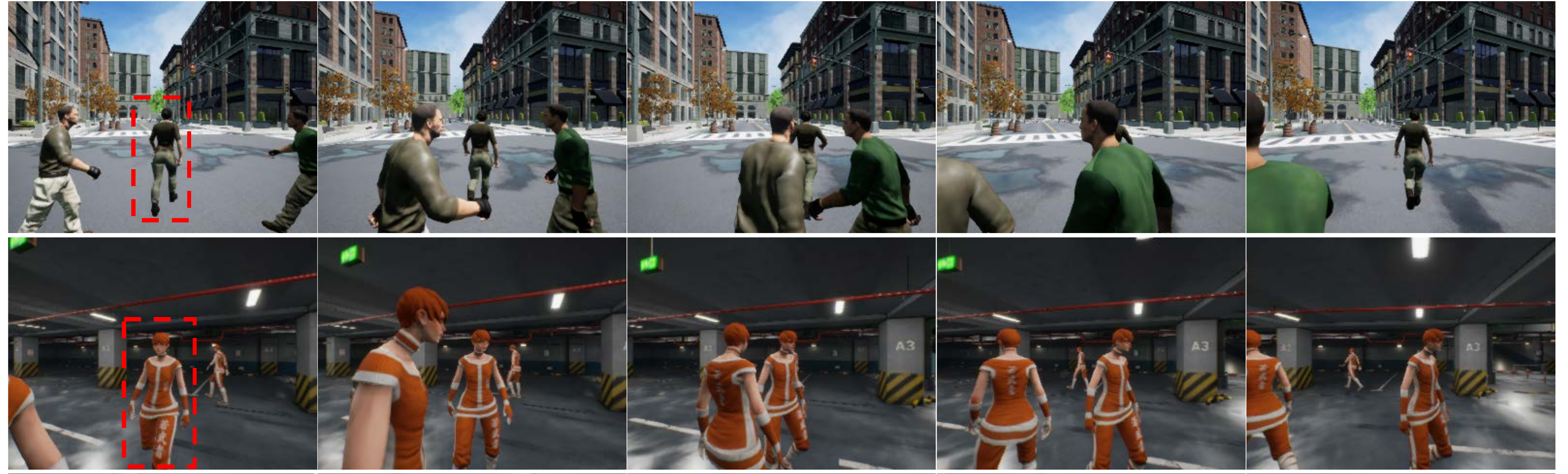}
\vspace{-0.3cm}
\caption{Two exemplar sequences of our tracker running on the \emph{Urban City} (top) and \emph{Parking Lot} (bottom). For better understanding, we point out the target object with a red line bounding box in the first frame.
More examples are available in the demo video.
}
\label{fig:demo_seq}
\end{figure}

As the complexity of the environment increases, performance of these models is downgraded comparing to the results in \emph{Simple Room}, shown as Fig.~\ref{fig:generalization}.
Even so, our tracker still significantly outperforms others, showing the stronger transferability of our model. 
In particular, in \emph{Parking Lot} where the target and distractor have the same appearance, the tracker must be able to consider the spatial-temporal consistency to identify the target.
Correspondingly, trackers which mainly rely on the difference in appearance is not capable of perceiving the consistency, there they should not perform well.
In contrast, our tracker performs well in such cases, from which we can infer that our tracker is able to learn spatial-temporal consistent representation that can be very useful when transferring to other environments.
Two typical sequences are shown in Fig.~\ref{fig:demo_seq}.

\section{Conclusion and Discussion}
Distracting objects are notorious for degrading the tracking performance. This paper offers a novel perspective on how to effectively train a distraction-robust active visual tracker, which is a problem setting that has barely been addressed in previous work. 
We propose a novel multi-agent game for learning and testing. Several practical techniques are introduced to further improve the learning efficiency, including designing reward function, two-stage teacher-student learning strategy, recurrent attention mechanism \etc. Empirical results on 3D environments verified that the learned tracker is more robust than baselines in the presence of distractors.

Considering a clear gap between our trackers and the ideal one (ours vs teacher model), there are many interesting future directions to be explored beyond our work.
For example, we can further explore a more suitable deep neural network for the visual tracker. 
For real-world deployment, it is also necessary to seek an unsupervised or self-supervised domain adaption method~\cite{hansen2021selfsupervised} to improve the adaptive ability of the tracker on novel scenarios. Besides, it is also feasible to extend our game on other settings or tasks, such as multi-camera object tracking~\cite{li2020pose}, target coverage problem~\cite{xu2020hitmac}, and moving object grasping~\cite{fang2019dher}.

\section*{Acknowledgements}
Fangwei Zhong, Tingyun Yan and Yizhou Wang were supported in part by the following grants: MOST-2018AAA0102004, NSFC-61625201, NSFC-62061136001, the NSFC/DFG Collaborative Research Centre SFB/TRR169 ``Crossmodal Learning" II, Tencent AI Lab Rhino-Bird Focused Research Program (JR201913), Qualcomm University Collaborative Research Program.


\bibliography{main}
\bibliographystyle{icml2021}
\clearpage
\twocolumn[
\icmltitle{Appendix}
]
\appendix
\section{Learning Meta Policies}
In this section, we introduce the implementation details of the meta policies.
\subsection{State Pre-Processing}
To make the neural network better use the grounded state, we pre-process $\rho$ and $\theta$ at first.
We transform the global state into the entity-centric coordinate system for each agent.
To be specific, the input of agent $i$ is a sequence about the relative poses between each agent, represented as ${P_{i,1}, P_{i,2}, ... P_{i,n}}$, where $n$ is the number of the agents and $P_{i.j}=(\rho_{i,j}, cos(\theta_{i,j}), sin(\theta_{i,j}), cos(\phi_{i,j}), sin(\phi_{i,j})$. Note that $(\rho_{i,j}, \theta_{i,j}, \phi_{i,j})$ indicates the relative distance, angle, and relative orientation from agent $i$ to agent $j$. Agent $i$ is at the origin of the coordination.

\subsection{Network Architecture}
\begin{figure}[h]
\centering
\hspace*{0.01\linewidth} \\
\includegraphics[width=0.98\linewidth]{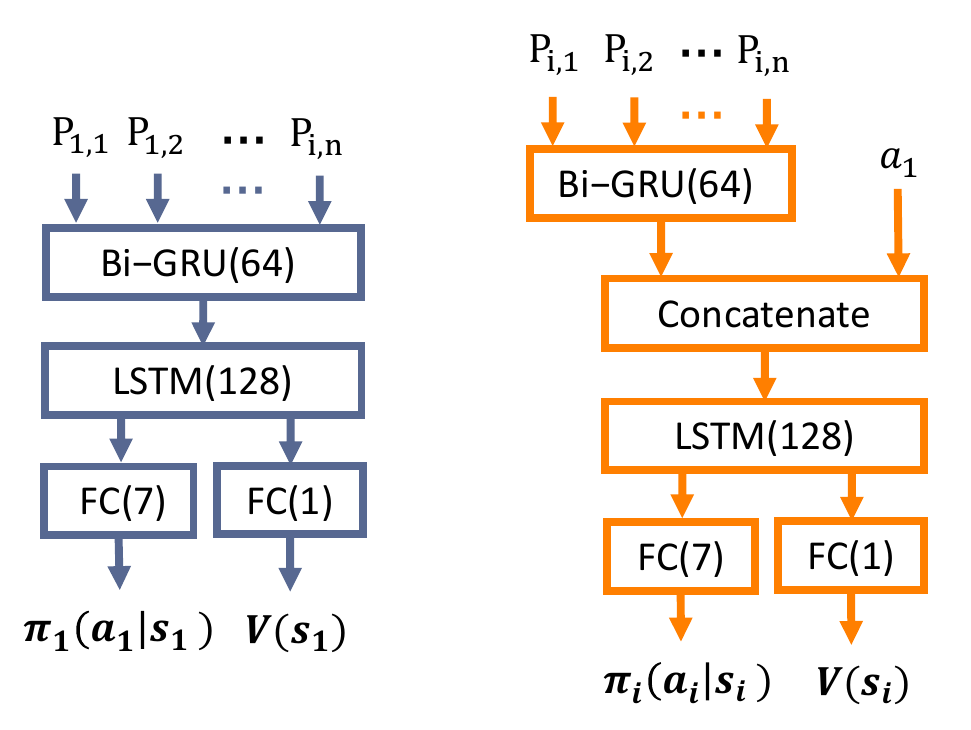}
\caption{ The network architectures for the meta policies. The left is the network for the tracker. The right is the network for the target and distractors.}
\label{fig:meta_net}
\end{figure}

 Since the number of the distractors is randomized during either training or testing, the length of the input sequence would be different across each episode.
 Thus, we adopt the Bidirectional-Gated Recurrent Unit (Bi-GRU) to pin-down a fixed-length feature vector, to enable the network to handle the variable-length distractors. 
 Inspired by the tracker-award model introduced in AD-VAT~\cite{zhong2018advat}, we additionally fed the tracker's action $a_1$ to the target and distractors, to induce a stronger adversarial policy.
 The network architectures are shown in Fig.~\ref{fig:meta_net}.
 Specifically, Bi-GRU(64) indicates that the size of the hidden state in Bi-GRU (Bi-directional Gated Recurrent Unit) is 64. Consequently, the size of the output of Bi-GRU is 128, two times of the hidden state. LSTM(128) indicates that the size of the hidden states in the LSTM (Long-Short Term Memofy) are 128.
 Following the LSTM, two branches of FC (Fully Connected Network) correspond to Actor(FC(7)) and Critic(FC(1)). 
 The FC(7) indicates a single-layer FC with 7 dimension output, which correspond to the 7-dim discrete action.
 The FC(1) indicates a single-layer FC with 1 dimension output, which correspond to the value function.
 The network parameters among the three agents are independent but shared among distractors.

\subsection{Training Details}
We adopt A3C~\cite{a3c-pytorch} to train the agent. We find that the reward normalization techniques can make the multi-agent learning process more efficient and steady. 
To improve the generalization of the policy for different numbers of distractors, we randomly set the number of distractors in the range of $0$ to $4$ for each episode.
The details of the hyper-parameters are introduced in Table.~\ref{meta-hyper}.

After that, we further finetune the meta tracker by playing against the saved models. Specifically, we randomly load the parameters from the model pool into the target and distractors networks at the beginning of each episode, and only the tracker network is optimized by RL during training. The details of the hyper-parameters are introduced in Table.~\ref{finetune-hyper}.
\begin{table}[tb]
  \caption{Hyper-parameters for learning meta policies}
  \label{meta-hyper}
  \centering
  \begin{center}
  \begin{tabular}{lc}
    \hline
    Hyper-parameters    &   \#   \\
    \hline
    \centering
  total steps     & 2M       \\
  episode length        & 500  \\
  discount factor       & 0.9     \\
  entropy weight for tracker       & 0.01    \\
    entropy weight for target and distractor       & 0.03    \\
  number of forward steps in A3C & 20 \\
  learning rate         & 1e-3 \\
  number of workers               & 4     \\
  optimizer          & Adam  \\
    \hline
  \end{tabular}
  \end{center}
\end{table}

\begin{table}[tb]
  \caption{Hyper-parameters for finetuning meta tracker}
  \label{finetune-hyper}
  \centering
  \begin{center}
  \begin{tabular}{lc}
    \hline
    Hyper-parameters    &   \#   \\
    \hline
    \centering
  total steps     & 0.5M       \\
  episode length        & 500  \\
  discount factor       & 0.9     \\
  entropy weight for tracker       & 0.01    \\
  entropy weight for target and distractor       & 0.03    \\
  number of forward steps in A3C & 20 \\
  learning rate         & 5e-4  \\
  number of workers               & 4     \\
  optimizer          & Adam  \\
    \hline
  \end{tabular}
  \end{center}
\end{table}

\section{Learning Active Visual Tracker}
\label{sec:student}
In this section, we introduce the implementation details of the active visual tracker.
\subsection{Network Architecture}
\begin{figure}[tb]
\centering
\hspace*{0.01\linewidth} \\
\includegraphics[width=0.98\linewidth]{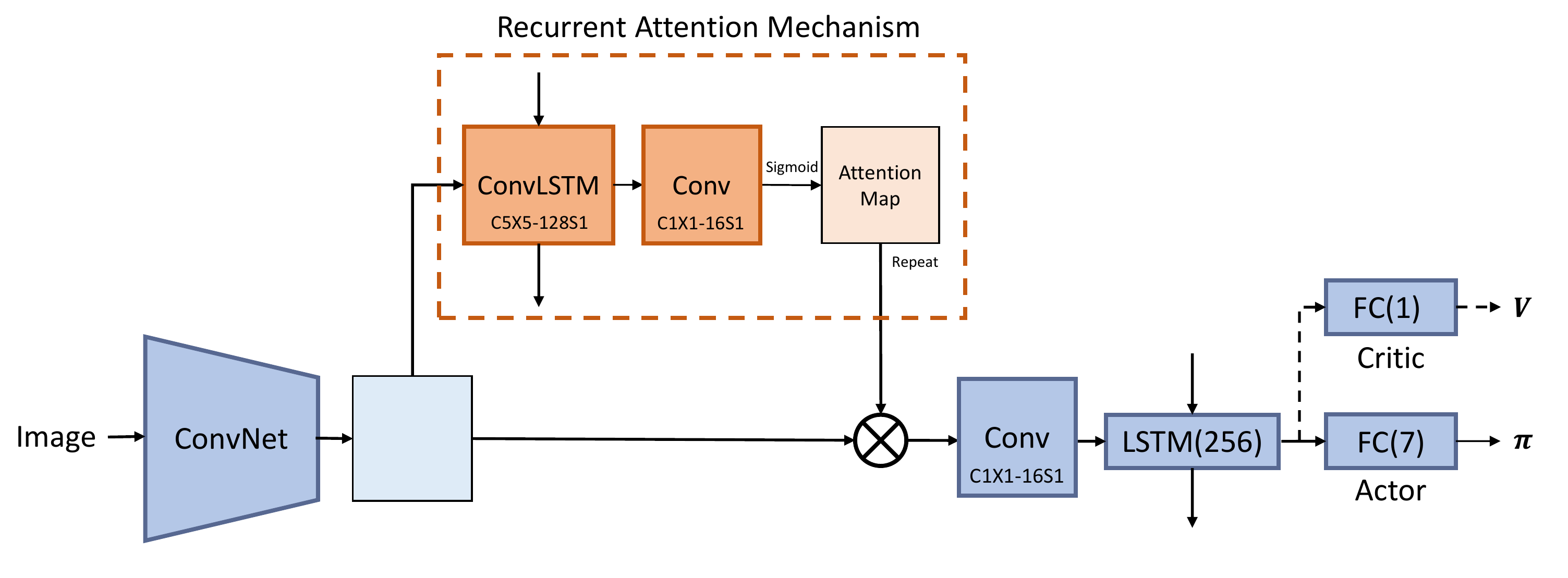}
\caption{The network architecture for active visual tracking. C1$\times$1-16S1 indicates a convolutional layer of 16 filters of size 1$\times$1 and stride 1.}
\label{fig:tracker_net}
\vspace{-0.3cm}
\end{figure}

The network for visual tracker is based on the Conv-LSTM network introduced in ~\cite{luo2019pami}.
The main difference to previous work~\cite{zhong2018advat, luo2019pami} is that we additionally adopt a recurrent attention branch to enhance the response of the target in the feature map, shown as Fig.~\ref{fig:tracker_net}.
In details, we employ the ConvLSTM~\cite{xingjian2015convolutional} to memorize the appearance information of the target by updating its cell $c_t$ and hidden states $h_t$.
Comparing to the conventional LSTM network, it can preserves the target’s
spatial structure, which is useful for localizing the target in the recent frame. 
Note that in the recurrent attention mechanism, Conv layer is followed by a hardsigmoid activation function. 

As for the overall network, the raw-pixel image observation is encoded by three layers Convolutional Neural Networks (ConvNet). 
 The specification of the ConvNet is given in the following table:
\begin{scriptsize}
\begin{center}
\resizebox{\linewidth}{!}{
\begin{tabular}{|c|c|c|c|c|c|c|c|}
\hline
Layer\# & 1 & 2 & 3 & 4 & 5 \\
\hline
ConvNet & C5$\times$5-32\emph{S}2 & MP 2\emph{S}2& C5$\times$5-64\emph{S}1 & MP 2\emph{S}2 &
C3$\times$3-128\emph{S}1 \\
\hline
\end{tabular}
}
\end{center}
\end{scriptsize}
 where C5$\times$5-32\emph{S}2 means 32 filters of size 5$\times$5 and stride 2,  MP 2\emph{S}2 indicates max pool of square window of size=2, stride=2.
The ConvNet are learned from scratch.
The input image is resized to $120 \times 120 \times 3$  RGB image.
Each layer is activated by a LeakyReLU function.
Layer \# 1 and \# 2 are followed by a $2\times2$ max-pooling layer with stride 2, additionally.
After that, the feature map is multiplied by the attention map.
Then we employ a convolutional layer and a LSTM(256) to furhter encode the state representation for the actor and critic.
Note that the critic is only used for the ablation (w/o teacher-student learning) that train the networks via reinforcement learning. In our teacher-student learning paradigm, the critic is not necessary at all.
As for the w/o recurrent attention, we just remove the recurrent attention branch and directly feed the feature map from ConvNet to the following.
The training process is presented in Algorithm. \ref{alg:example}.

\subsection{Training Details}
The details of the hyper-parameters are introduced in Table.~\ref{ts-hyper}.
The details of the teacher-student learning process are introduced in Algorithm.~\ref{alg:example}
\begin{table}[tb]
  \caption{Hyper-parameters for our teacher-student learning method.}
  \label{ts-hyper}
  \centering
  \begin{center}
  \begin{tabular}{lc}
    \hline
    Hyper-parameters    &   \#   \\
    \hline
    \centering
  total steps     & 2M       \\
  max episode length        & 500  \\
  buffer size & 500 episodes \\
  batch size &  8 \\
  number of forward steps & 30 \\
  learning rate         & 1e-4  \\
  number of samplers               & 4     \\
  optimizer          & Adam  \\
    \hline
  \end{tabular}
  \end{center}
\end{table}

\begin{algorithm}[tb]
  \caption{Teacher-Student Learning}
  \label{alg:example}
\begin{algorithmic}
  \STATE {\bfseries Require:} meta tracker $\pi_1^*$, a model pool for targets $\Pi_2$ and distractors $ \Pi_3$ 
  \STATE {\bfseries Initialize:} Randomly initialize student policy $\pi_1$, Replay Buffer $B$\\
  \% sample interactions, can run with multiple workers.
  \FOR{simulation episode $e=1$ {\bfseries to} $M$}
  \STATE sample a target policy $\pi_2$ from $\Pi_2$.
  \STATE sample a distractor policy $\pi_3$ from $\Pi_3$.
  \FOR{$t=0$ {\bfseries to} $T$}
  \STATE obtain visual observation for tracker $o_{1,t}$
  \STATE obtain grounded state for each agent $s_{1,t},s_{2,t},..s_{n,t}$
  \STATE Execute actions $a_{1,t} = \pi_1(o_{1,t})$, $a_{2,t} = \pi_1(s_{2,t})$, $a_{3,t} = \pi_1(s_{3,t})$
  \STATE Compute suggestions for tracker $a_1^* = \pi_1^*(s_{1,t}) $
  \STATE Store $(o_{1,t}, a_1^*)$ in buffer $B$
  \ENDFOR\\
  \% Asynchronously update student network
  \WHILE{$e < M$}
    \STATE Sample a batch of $N$ sequences from buffer $B$ 
    \STATE Train student policy ($\pi_1(o_t)$) by optimizing $\mathcal{L}_{KL}$.\\
  \ENDWHILE
  \ENDFOR
\end{algorithmic}
\end{algorithm}

\section{Baselines}
In this section, we introduce the details of baselines used in the experiments.
\subsection{Two-Stage Methods}
In the first stage, we extract the bounding box of the target from image by passive (video-based) visual object tracker.
Then, to meet the requirements of active tracking, we build a controller based on the output bounding box additionally.

\textbf{Visual Tracker.}
In this paper, we adopt three off-the-shelf trackers as our baselines, e.g. DaSiamRPN~\cite{zhu2018distractor}, ATOM~\cite{danelljan2019atom}, DiMP~\cite{bhat2019learning}. 
We implement them based on their official repository.
Specifically, DaSiamRPN is from \url{https://github.com/foolwood/DaSiamRPN}, ATOM and DiMP are from \url{https://github.com/visionml/pytracking}

\begin{figure}[t]
\begin{center}
\includegraphics[width=0.85\linewidth]{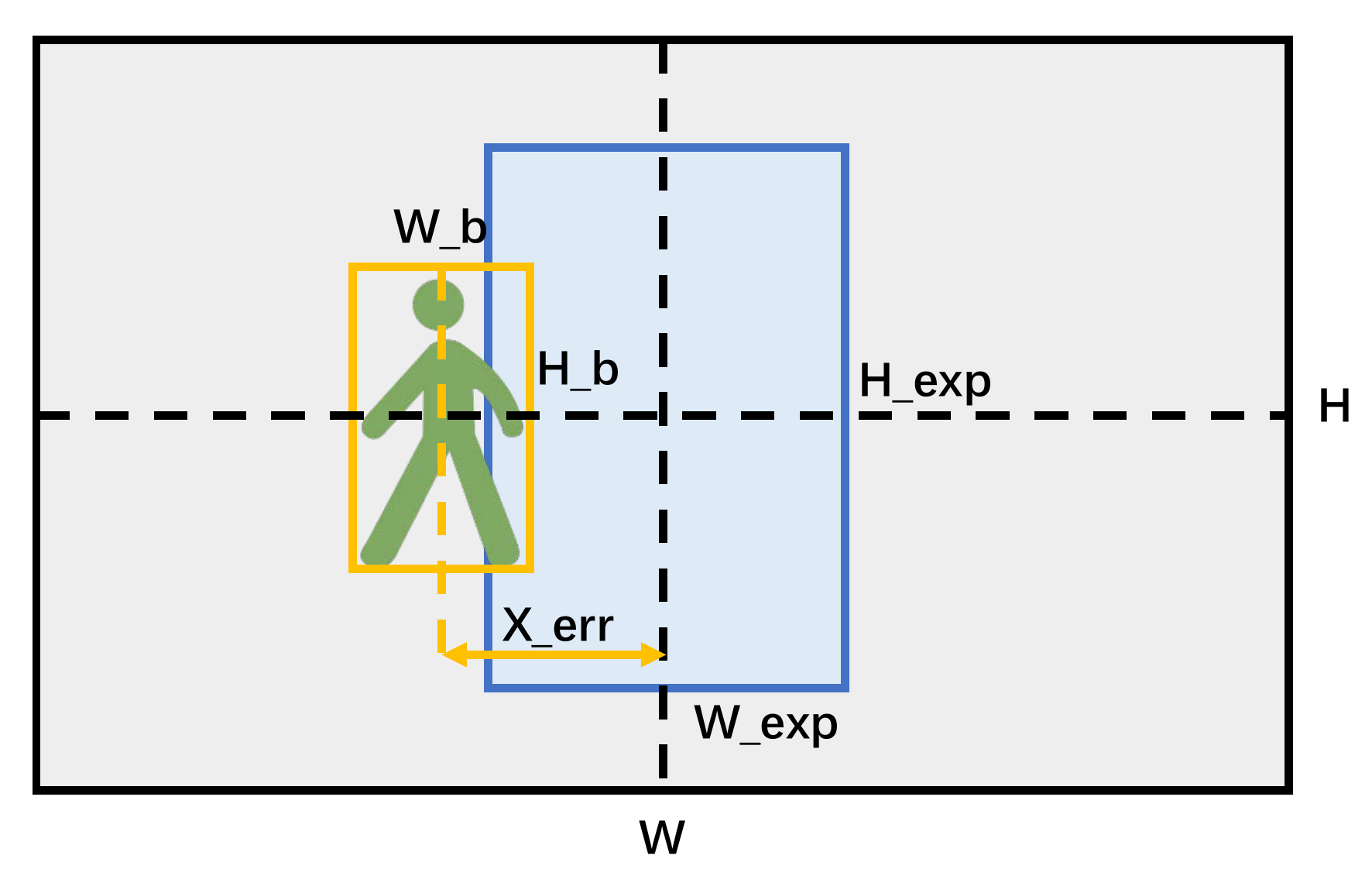}
\end{center}
\vspace{-0.4cm}
\caption{An example to illustrate errors.}
\vspace{-0.3cm}
\label{fig:pid}
\end{figure}

\textbf{Camera Controller.}
The camera controller outputs actions based on the specific error of the bounding box.
The controller choose a discrete action from the action space according to the horizontal error $X_{err}$ and the size error $S_{err} = W_b\times H_b - W_{exp}\times H_{exp}$, shown as Fig.~\ref{fig:pid}

Intuitively, $X_{err}$ measures how far the bounding box is horizontally away from the desired position. In the case of Fig. \ref{fig:pid}, $X_{err}$ is negative, meaning that the camera should move left so the object is closer to the center of the image.
The distance to the target is reflected by the size of the bounding box $S_{err}$.
Intuitively, we have the following commands,

$X_{err} < 0$ means that the tracker is to the left and needs to turn right (increasing $X$).

$X_{err} > 0$ means that the tracker is to the right and needs to turn right (decreasing $X$).

$S_{err} < 0$ means that the tracker is too far away and needs to move forward and speed up (increasing $S$).

$S_{err} > 0$ means that the tracker is too close and needs to move backward or slow down (decreasing $S$).

Formally, the output signal in the linear velocity $V_l$ and angular velocity $V_a$ is depended on $S_{err}$ and $X_{err}$ respectively, $V_l= - P_l * S_{err}$, $V_a = - P_a * X_{err}$. Note that $P_l>0$ and $P_a>0$ are two constant, the higher the more sensitive to the error. Note that, to keep the comparison fair, we discretize the output to map it to the discrete action space we used in other methods.
\subsection{End-to-End Methods}
There are four end-to-end methods (SARL~\cite{luo2019pami}, AD-VAT~\cite{zhong2018advat}, and their variants) are employed as the baselines.
We implement them based on the official repository in \url{https://github.com/zfw1226/active_tracking_rl}.

\textbf{Network Architecture.}
The network architecture follows the networks used in ~\cite{zhong2018advat}.
Different to original implementation, we modify the network in two aspects: 1) use color image as input, instead of the gray image used in previous to learn a more discriminative representation. 2) add an auxiliary task for tracker, i.e. predict its immediate reward, to speed up the learning of tracker. 

\textbf{Optimization}.
The network parameters are updated with a shared Adam optimizer.
Each agent is trained by A3C \cite{mnih2016asynchronous}.
$4$ workers are running in parallel during training. 
The hyper-parameters we used is the same as the original version. 

\section{Environments}
In this section, we introduce the details of the training and testing environments used in the experiments.
\subsection{Training Environment}
We train the agents in the \emph{Simple Room}. 
The environment augmentation technique\cite{luo2019pami,zhong2018advat} is employed to randomize the illumination and visual appearance of objects (target, distractors, backgrounds).
When learning the meta policies, the number of distractors is randomized, ranging from 0 to 4.
But the number is fixed at 2, when learning the visual policies.
Environment augmentation can significantly improve the generalization of the visual policy.
To produce more challenging and realistic setting, we modify the action space in two ways: 1) increase the max speed of players from 1m/s to 2m/s, 2) add a filter $v_ t = \alpha v_{t-1} + (1-\alpha) v_t$ to smooth the discrete action.

\subsection{Testing Environment}
\begin{figure}[tb]
\centering
\hspace*{0.01\linewidth} \\
\includegraphics[width=0.19\linewidth]{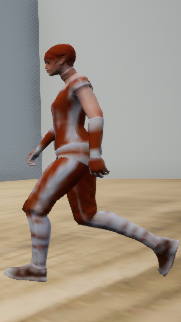}
\includegraphics[width=0.19\linewidth]{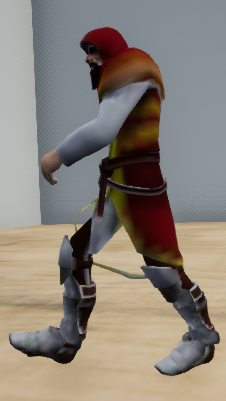}
\includegraphics[width=0.19\linewidth]{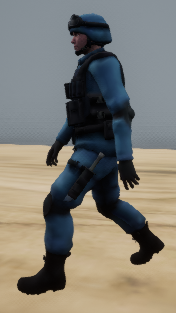}
\includegraphics[width=0.19\linewidth]{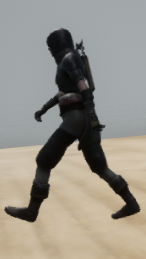}
\includegraphics[width=0.19\linewidth]{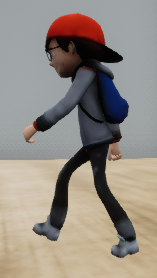}
\includegraphics[width=0.24\linewidth]{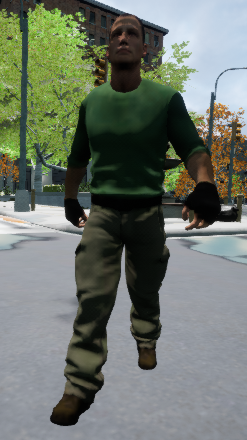}
\includegraphics[width=0.24\linewidth]{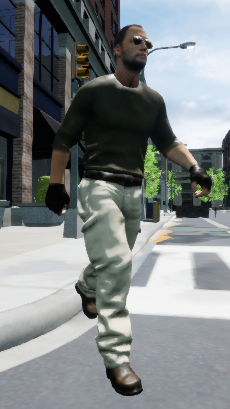}
\includegraphics[width=0.24\linewidth]{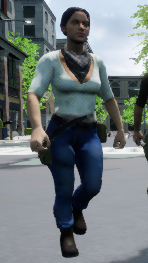}
\includegraphics[width=0.24\linewidth]{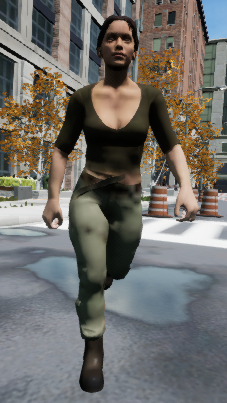}
\caption{\textbf{Top:} Five targets appear in \emph{Simple Room}. The one on the leftmost is also used in \emph{Parking Lot}. \textbf{Bottom:} Four candidate appearances used for target and distractors in \emph{Urban City}, including two man and two woman.
}
\label{fig:room}
\end{figure}

\begin{figure}[tb]
\centering
\includegraphics[width=0.32\linewidth]{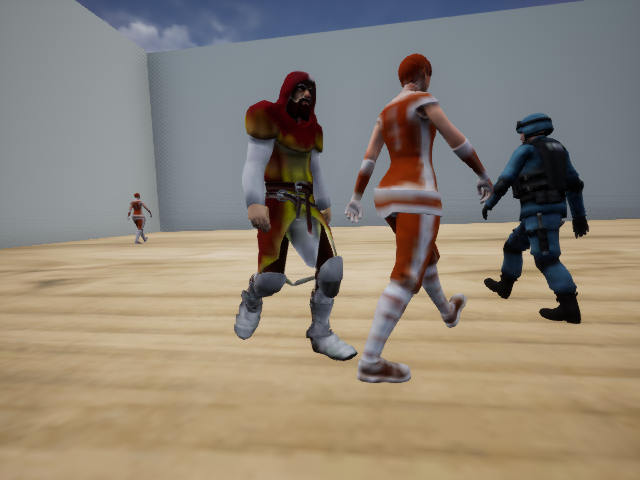}
\includegraphics[width=0.32\linewidth]{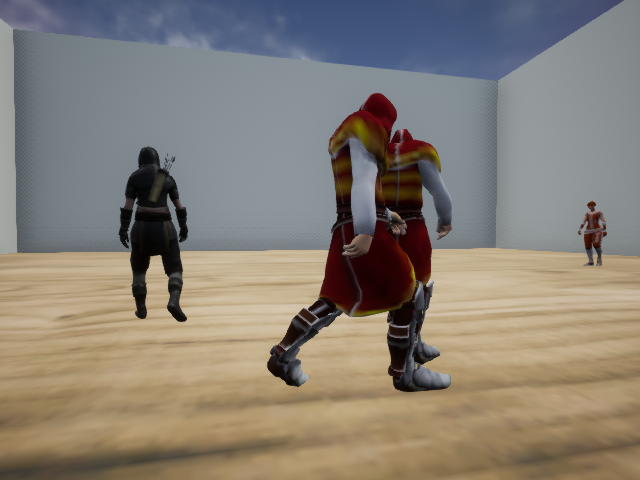}
\includegraphics[width=0.32\linewidth]{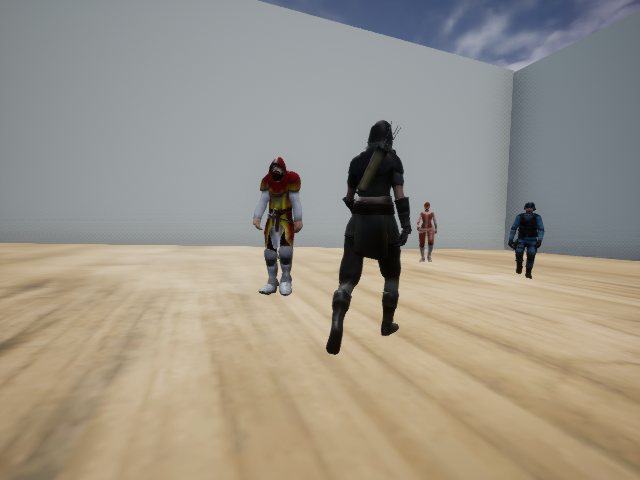}
\includegraphics[width=0.32\linewidth]{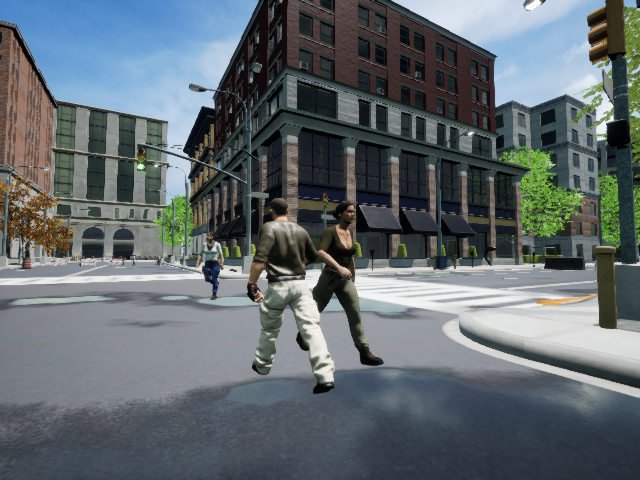}
\includegraphics[width=0.32\linewidth]{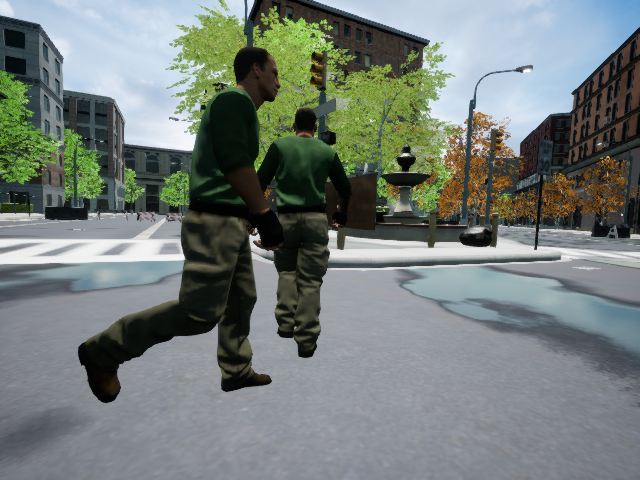}
\includegraphics[width=0.32\linewidth]{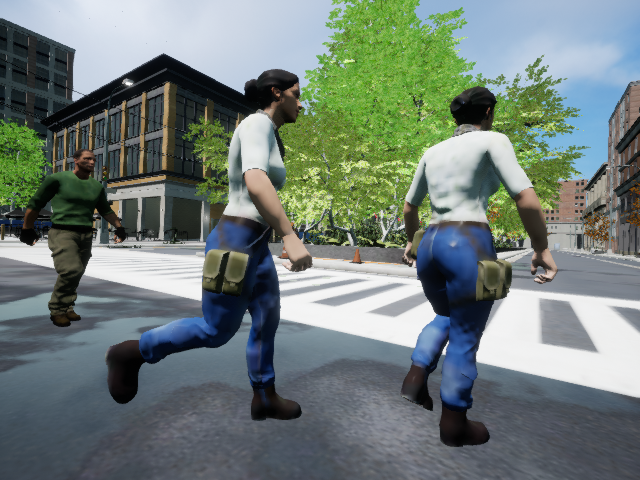}
\includegraphics[width=0.32\linewidth]{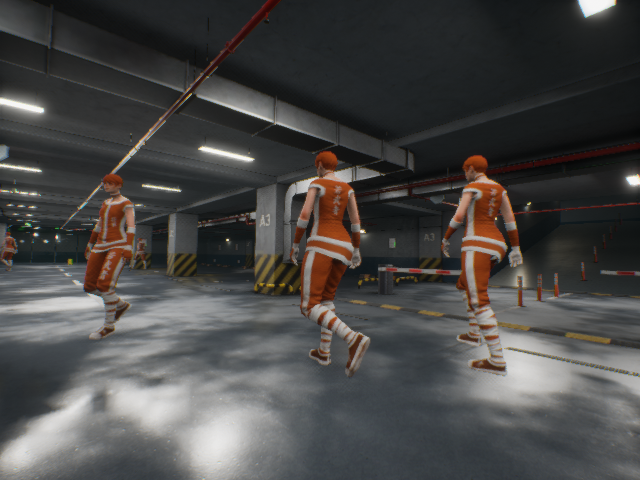}
\includegraphics[width=0.32\linewidth]{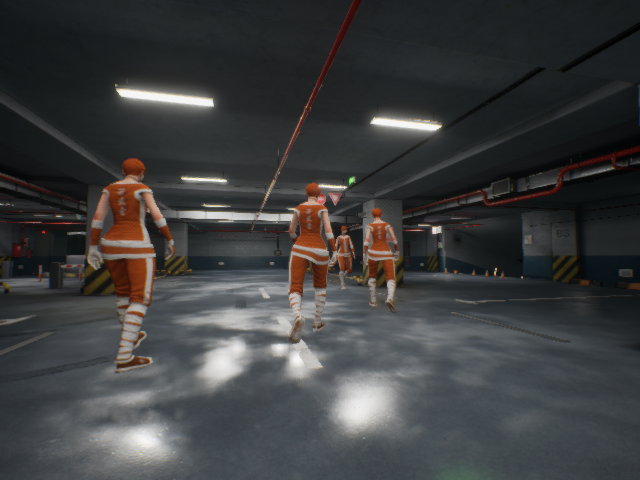}
\includegraphics[width=0.32\linewidth]{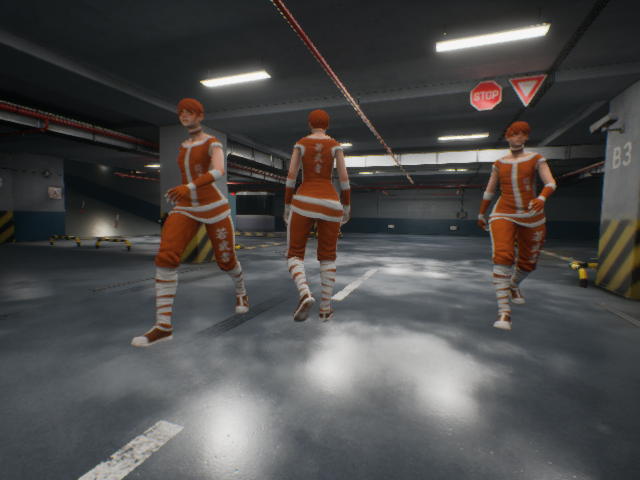}
\caption{
Typical examples of the tracker's first-person view in the testing environments. From top to bottom are \emph{Simple Room}, \emph{Urban City}, and \emph{Parking Lot} }
\label{fig:example}
\end{figure}

There are three environments used for testing, \ie \emph{Simple Room}, \emph{Urban City}, and \emph{Parking Lot}.
There are five appearances used in the \emph{Simple Room}, shown as the top of Fig.\ref{fig:room}.
There are four appearances used in the \emph{Urban City}, shown as the bottom of Fig.\ref{fig:room}.
Since the dress of the target and distractors are randomly sampled at each episode, two players can be dressed the same.
In \emph{Parking Lot}, all the target and distractors are of the same dress, shown as the bottom of Fig.\ref{fig:example}.
All appearances used by the target and distractors in the testing environments are different from the training environment.
Graphics in Fig.\ref{fig:example} are examples of the three testing environment.

\begin{figure}[tb]
\centering
\hspace*{0.01\linewidth} \\
\includegraphics[width=0.99\linewidth]{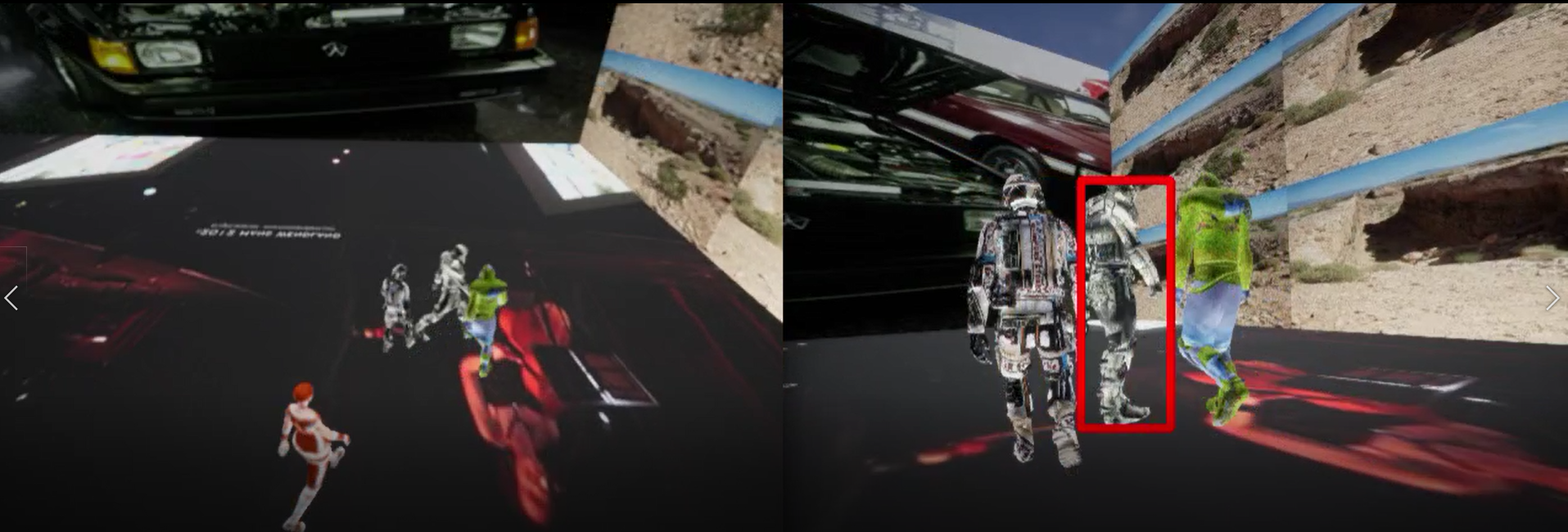}
\includegraphics[width=0.99\linewidth]{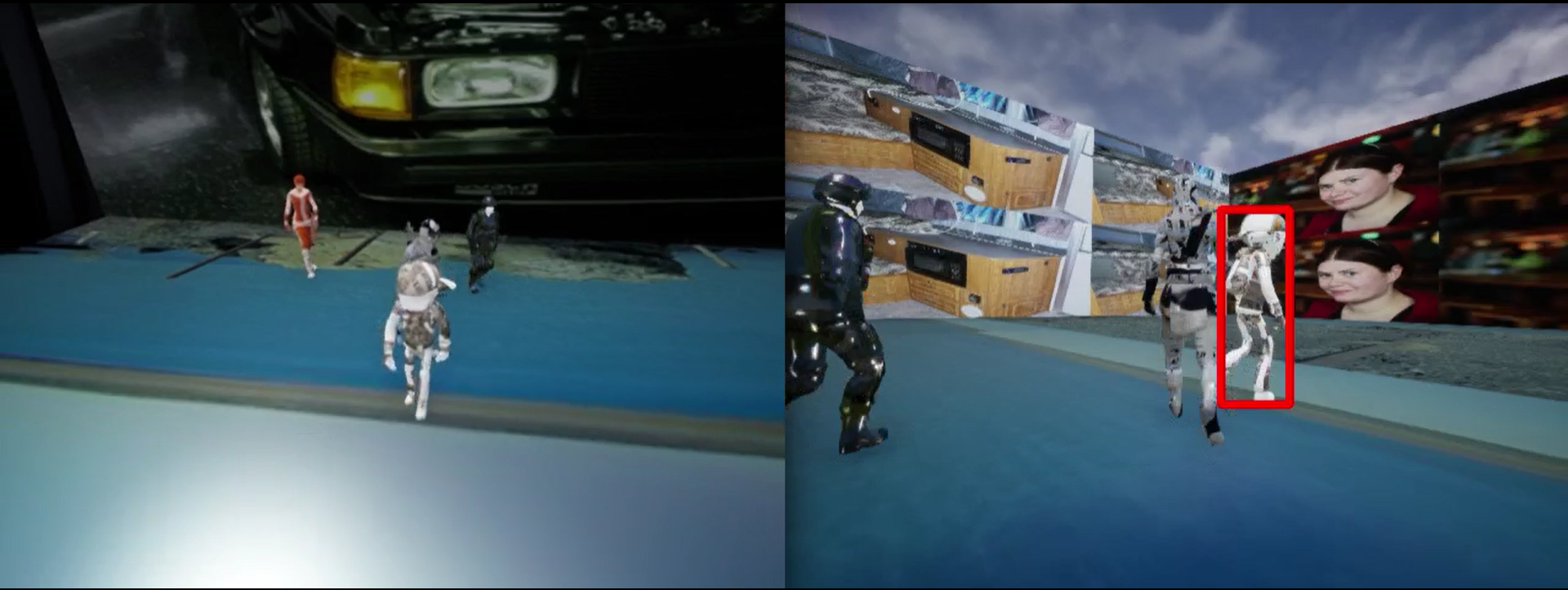}
\includegraphics[width=0.99\linewidth]{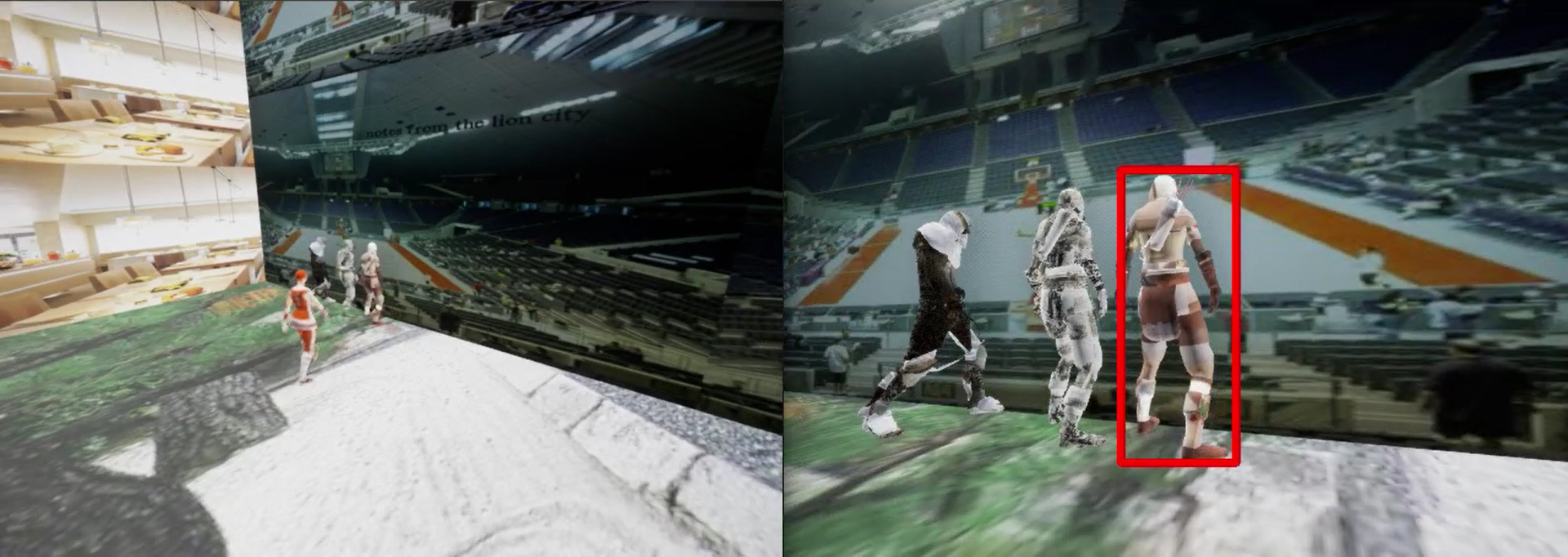}
\caption{Three snapshots of the training environment. The left is captured from the bird view. The right is the visual observation for the tracker. The target is noted by the red bounding box in the tracker's view. }
\label{fig:aug}
\end{figure}

\section{Detailed Results and Demo Videos}
In this section, we report detailed results with the Accumulated Reward (AR), Episode Length (EL), and Success Rate (SR), and show demo videos for better understanding.

\subsection{Multi-Agent Curriculum}
In Video 1, we show the behaviors of the target and distractors at different learning stages when learning in the multi-agent game. Note that the tracker is governed by the learned meta tracker for better visualization.

\subsection{Evaluating with Scripted Distractors}
See Table.\ref{3d-table} for quantitative results.
\begin{table*}[tbh]
\caption{Evaluating the active visual trackers on \emph{Simple Room} with different number of scripted distractors.}
\resizebox{\textwidth}{!}{
\begin{tabular}{l|lll|lll|lll|lll|lll}
\hline
\centering
  Methods        & \multicolumn{3}{c|}{Nav-0} & \multicolumn{3}{c|}{Nav-1} & \multicolumn{3}{c|}{Nav-2} & \multicolumn{3}{c|}{Nav-3} & \multicolumn{3}{c}{Nav-4} \\
          & AR     & EL     & SR      & AR     & EL     & SR      & AR     & EL     & SR    & AR     & EL     & SR      & AR     & EL     & SR      \\
\hline
DaSiamRPN & 170    & 460    & 0.80    & 58     & 343    & 0.45    & 45     & 321    & 0.37    & 15     & 302    & 0.27    & 3      & 259    & 0.23    \\
ATOM      & 286    & 497    & 0.94    & 210    & 423    & 0.77    & 193    & 403    & 0.63    & 117    & 337    & 0.40   & 103    & 318    & 0.38     \\
DiMP      & 336    & \bf500    & \bf1.00    & 255    & 449    & 0.76    & 204    & 399    & 0.59    & 113    & 339    & 0.38    & 97    & 307    & 0.26    \\
SARL      & 368    & \bf500    & \bf1.00    & 244    & 422    & 0.64    & 163     & 353    & 0.46    & 112     & 325    & 0.36    & 87     & 290    & 0.23    \\
SARL+      & 373    & \bf500    & \bf1.00    & 226    & 407    & 0.60    & 175     & 370    & 0.52    & 126     & 323    & 0.38    & 22     & 263    & 0.15    \\
AD-VAT    & 356    & \bf500    & \bf1.00    & 221    & 401    & 0.62  &  171     & 376    & 0.50    & 52     & 276    & 0.18     & 16     & 223    & 0.16    \\
AD-VAT+    & 373    & \bf500    & \bf1.00    & 220    & 418    & 0.64    & 128     & 328    & 0.35    & 96     & 304    & 0.34     & 35     & 262     & 0.18    \\
\hline
Ours      & \bf412    & \bf500    & \bf1.00    & \bf357    & \bf471    & \bf0.88    & \bf303    & \bf438    & \bf0.76    & \bf276    & \bf427    &  \bf0.65    & \bf250    & \bf401    & \bf0.54    \\
\hline
Teacher   & 420    & 500    & 1.00    & 413    & 500    & 1.00    & 410    & 500    & 1.00    & 409    & 500    & 1.00    & 407    & 500    & 1.00   \\
\hline
\end{tabular}
}
\label{3d-table}
\end{table*}

\subsection{Adversarial Testing}
In Video 2, we show the emergent target-distractor cooperation after learning to attack DiMP, ATOM, AD-VAT, and Ours.

\subsection{Transferring to Realistic Virtual Environments}
See Table.\ref{transfer} for quantitative results. 
More vivid examples for our tracker are available in Video 3.
Note that the videos are recorded from the tracker's view.
\begin{table}[tbh]
\caption{Comparison with baselines in unseen environment. Note that there are four  distractors in \emph{Urban City} and two distractors in \emph{Parking Lot}. The best results are shown in bold.}
\label{transfer}
\begin{tabular}{l|lll|lll}
\hline
\centering
Methods & \multicolumn{3}{c|}{\emph{Urban City}} & \multicolumn{3}{c}{\emph{Parking Lot}} \\
 & AR       & EL       & SR        & AR       & EL       & SR        \\
\hline
DaSiamRPN                                   & 120      & 301      & 0.24      & 52       & 195      & 0.10       \\
ATOM                                         & 156      & 336      & 0.32      & 113      & 286      & 0.25      \\
DiMP                                         & 170      & 348      & 0.38     & 111      & 271     & 0.24      \\
SARL                                         & 97      & 254      & 0.20       & 79       & 266      & 0.15      \\
SARL+                                         & 74      & 221      & 0.16       & 53       & 237      & 0.12      \\
AD-VAT                                       & 32       & 204      & 0.06      & 43       & 232      & 0.13      \\
AD-VAT+                                       & 89       & 245      & 0.11      & 35       & 166      & 0.08      \\
\hline
Ours & \bf227      & \bf381      & \bf0.51      & \bf186      & \bf331      & \bf0.39     \\
\hline
\end{tabular}
\end{table}

\end{document}